\documentclass{article}

\usepackage[nonatbib,final,main]{neurips_2026}

\usepackage{amsmath}
\usepackage{amssymb}
\usepackage{siunitx}

\usepackage[
  style=authoryear-comp,
  maxcitenames=1,
  natbib=true,
  uniquename=false,
  uniquelist=false,
  date=year,
  backref=true,
]{biblatex}
\AtEveryBibitem{
  \iffieldundef{doi}{}{ % remove other links if there is a DOI
    \clearfield{url}
    \clearfield{eprint}
  }
  \iffieldundef{eprint}{}{ % remove url if there is an eprint
    \clearfield{url}
  }
  \clearfield{language}
  \clearfield{addendum}
  \clearfield{urldate}
}
\usepackage{listings}

\usepackage[algoruled,vlined]{algorithm2e}

\usepackage{enumitem}

\usepackage{hyperref}

\usepackage{booktabs}

\usepackage{graphicx}
\usepackage{tikz}
\usetikzlibrary{positioning,fit,shapes.geometric}
\newcommand{\includempl}[1]{\includegraphics[scale=0.703125]{#1}}
\newcommand{\widecenter}[1]{\noindent\hspace{-\textwidth}\makebox[3\textwidth][c]{#1}}

\usepackage[disable,textsize=tiny]{todonotes}
\newcommand{\redactedcite}[1]{#1}

\title{Very Fast Bayesian Additive Regression Trees on GPU}

\author{%
  Giacomo Petrillo\\
  Independent, formerly Department of Statistics, Computer Science,\\
  Applications ``Giuseppe Parenti'' (DISIA), University of Florence, Italy\\
  \texttt{giacomo.petrillo@unifi.it}\\
}

\begin{document}

\maketitle

\begin{abstract}

	Bayesian Additive Regression Trees (BART) is a nonparametric Bayesian
	regression technique based on an ensemble of decision trees. It is part
	of the toolbox of many statisticians. The overall statistical quality
	of the regression is typically higher than other generic alternatives,
	and it requires less manual tuning, making it a good default choice.
	However, it is a niche method compared to natural competitors such as
	XGBoost, due to the longer running time, making sample sizes
	above \num{10000}--\num{100000} a nuisance. We present a GPU
	implementation of BART, up to 500x faster than the previous state of
	the art, making BART fast enough to be a viable option on large
	datasets, and showing that BART parallelizes better than its
	non-Bayesian counterparts. We also find that, to be as practical as
	tree boosting on large datasets, BART would need a scalable way to
	infer the number of trees, which we do not provide. Our implementation is available in the
	Python package \texttt{bartz}.

\end{abstract}

\section{Introduction}
\label{sec:intro}

Bayesian Additive Regression Trees (BART) is a nonparametric Bayesian regression
method, introduced by \citet{chipman2006,chipman2010}. It defines a prior
distribution over the space of functions by representing them as a sum of binary
decision trees, and then specifying a stochastic tree generation process. The
posterior is then obtained with Metropolis-Gibbs sampling over the trees. BART
can be considered as a Bayesian version of tree boosting. See \citet{hill2020}
for a review, and \citet[ch.~5]{daniels2023} for a textbook treatment.

BART is empirically effective and gained some popularity \citep[consider,
	e.g.,][]{tan2019}. The Atlantic Causal Inference Conference (ACIC) Data
Challenge confirmed BART as one of the best regression methods for causal
inference \citep{dorie2019,hahn2019,acic2019,thal2023}. A good part of the BART
literature focuses on adding features to the core algorithm, e.g., missing
predictors \citep{kapelner_prediction_2015}, variable selection
\citep{linero2018}, smooth prediction \citep{linero2018b}, discrete and
categorical outcomes \citep{murray_log-linear_2021}, dynamic number of trees
\citep[ch.~3]{collins_improved_2023}, categorical predictors
\citep{deshpande_flexbart_2025}, and arbitrary conditional response distribution
\autocite[i.e., arbitrary ``likelihood'' or ``loss'',][]{linero_generalized_2025}.
Other works design models built on top of BART to handle specific needs, e.g.,
causal inference \citep{hahn2020}, missing data problems \citep{goh_joint_2024},
and time series \citep{huber_inference_2022,lima_minnesota_2025}. Some effort
was directed at computational optimization: data parallelism
\citep{pratola2014}, efficient implementations
\citep{dorie2024,deshpande_flexbart_2025}, and Accelerated BART
\autocite[XBART,][]{he2019,he_stochastic_2023}.

The natural non-Bayesian competitors to compare BART against are popular
gradient-boosted trees implementations like XGBoost \citep{chen2016}, LightGBM
\citep{ke_lightgbm_2017} and CatBoost \citep{prokhorenkova_catboost_2018}.
BART is known to achieve comparable performance to XGBoost with less tuning
\autocites[or better performance with comparable tuning, etc., see][table~2]{linero2018b}[supplementary]{he2019}[\S A]{nguyen2025oblique}. 
Furthermore, as a
Bayesian method, BART includes out-of-the-box uncertainty quantification, and,
as illustrated by the many variants in the literature, it can be extended to
larger models in a principled way, making BART a solid building block. Despite
these positive factors, BART is a niche method compared to its non-Bayesian
counterparts. According to the \citet[p.~35]{kaggle2021} survey, the most
popular regression tool amongst the ones mentioned is XGBoost, used by half of
the surveyed data scientists. From download counts, we estimate that BART is
$O(1000)$ times less used than XGBoost.

We hypothesize two reasons for this gap.
First, speed: a single BART run produces an ensemble of $\mathord{\sim}
	1000$ forests, representing the posterior distribution. Accordingly, BART is
$O(1000)$ times slower. The two largest BART analyses we know of had
$n\approx\num{500000}$ observations/examples and $p=90$ predictors/features
\citep{pratola2020}, and $n\approx\num{7000000}$, $p=3$ \autocite[\S7.2]{pratola2014}, while XGBoost boasts ``billions of examples'' on
its home page and was presented with a 1.7B$\times$67 dataset \autocite[table~2]{chen2016}.
Second, software ecosystem: for example, \citet{he_stochastic_2023} showed
that XBART was competitive in speed and accuracy with cross-validated XGBoost and Random Forests, but then their
software implementation was not maintained to the quality standards of popular
data science packages, with only recent efforts at providing a well-maintained
XBART implementation \citep{herren_stochtree_2026}.

This paper addresses the speed gap by re-implementing BART on GPU.
Our plain BART implementation is 100 times faster on GPU
vs.\ CPU (at $n \gtrsim \num{10000000}$, on an Nvidia RTX PRO 5000 Blackwell vs.\
an Apple M1 Pro), while also surpassing the previous state of the art
\autocite{dorie2024} by 5x on CPU, for a total 500x speed-up at large $n$. This
makes BART fast enough to be a (still generally slower) alternative to
non-Bayesian methods on large datasets,
with the inconvenience of requiring a GPU.\footnote{In a previous
	version of this manuscript, we claimed running time ``competitive with''
XGBoost, and a 200x speed-up vs.\ CPU. We moderated these claims as our CPU implementation improved more than our GPU one in the intervening time before this revision, and we fixed issues with the way we benchmarked XGBoost.}
While XGBoost
\& co.{} have their own GPU implementations as well, we find that our BART implementation gains more from GPU than XGBoost.
We provide a Python package,
\texttt{bartz} \redactedcite{\citep{gattocrucco_2026_23932737}}. In
\autoref{sec:bart} we summarize the background knowledge on BART, then in
\autoref{sec:impl} we describe how to adapt BART to run on GPU, and in
\autoref{sec:perf} we measure the performance of our implementation. Finally,
in \autoref{sec:conclusions} we assess the merits of our work and consider
further research directions.

\section{BART recap}
\label{sec:bart}

BART is a Bayesian nonparametric regression method. Semi-formally, this means
that, given a set of $n$ pairs of observations $(\mathbf x_i,y_i) \in \mathbb
R^p \times \mathbb R$, it infers without guidance a function $f$ such that $y_i
= f(\mathbf x_i) + \varepsilon_i$, keeping a balance between $f$ and the error
terms $\varepsilon_i$ such that given a new $\mathbf x^*$ from a pair $(\mathbf
x^*, y^*)$ with $y^*$ yet unknown, $f(\mathbf x^*)$ is a good advance
prediction of $E[y^*]$. The result is provided in the form of a set of possible
functions $\{f_k\}$, chosen in such a way that the spread of $f_k(\mathbf x^*)$
over the possibilities quantifies the uncertainty in predicting $E[y^*]$. More
precisely, this set is obtained by defining a prior distribution over the
space of functions, and then sampling the corresponding posterior distribution.
See \citet[ch.~4]{muller2015} for a general reference on this topic.

BART represents the regression function as the sum of a fixed number
$n_\text{tree}$ of decision trees. The prior on functions is induced by
defining a single-tree prior recursively, which is then repeated i.i.d.\ for
all trees. The posterior is sampled by a BART-specific MCMC algorithm that
updates one tree at a time. The implementations of BART most used in published research are the R
packages \texttt{bartMachine} \citep{kapelner2016,kapelner2023},
\texttt{dbarts} \citep{dorie2024}, and \texttt{BART}
\citep{sparapani2021,mcculloch2024}, though many others exist. In the
following, we summarize the description from the main BART paper
\citep{chipman2010}.

\paragraph{Decision trees}

A decision tree is a way to represent a step function $\mathbb R^p \to \mathbb R$
using a binary tree. Each internal (i.e., non-leaf) node contains a \emph{decision rule}, consisting of a
pair $(i, s)$ combining a splitting axis $i$ with a splitting point $s$ along such axis,
partitioning the input space into two half-spaces along the hyperplane $\{\mathbf x \in \mathbb
	R^p: x_i = s\}$, orthogonal to axis $i$. Each half is assigned to a child of the node.
Child nodes in turn further divide their space amongst their own children. Leaf nodes
contain the constant function value to be associated with their whole assigned rectangular
subregion. A minimal tree consisting of a single root leaf node represents a flat
function. See \autoref{fig:tree} for an illustration.

\begin{figure}
	\centering
	% Decision tree of Figure 1, redrawn after Hill, Linero & Murray (2020), Fig. 1.
% Requires \usetikzlibrary{positioning,fit,shapes.geometric}.
\begin{tikzpicture}[
		font=\small,
		% the parts of a decision rule are separate nodes, without padding, so
		% that the arrows can point at each part; they are aligned on the
		% baseline with the spacing of a relation symbol (5mu at \small)
		rulepart/.style={inner sep=0pt},
		relsep/.style={anchor=base west, xshift=2.5pt},
		decision/.style={draw, rectangle, inner xsep=8pt, inner ysep=6pt},
		leaf/.style={draw, circle, minimum size=7mm, inner sep=0pt},
		leafvalue/.style={inner sep=0pt},
		branch/.style={font=\sffamily\tiny},
		ann/.style={text=red!75!black, font=\sffamily\footnotesize, align=center},
		annarrow/.style={->, >=stealth, red!75!black, semithick},
		region/.style={draw, thin},
	]
	\colorlet{leafone}{blue!18}
	\colorlet{leaftwo}{orange!30}
	\colorlet{leafthree}{green!22}

	% the tree; the YES branch goes left, as in Figure 2

	\node[rulepart] (rax) at (2.6, 0) {$x_1$};
	\node[rulepart, relsep] (rlt) at (rax.base east) {$<$};
	\node[rulepart, relsep] (rsp) at (rlt.base east) {$0.8$};
	\node[decision, fit=(rax)(rsp)] (root) {};

	\node[rulepart] (iax) at (0.8, -1.05) {$x_2$};
	\node[rulepart, relsep] (ilt) at (iax.base east) {$<$};
	\node[rulepart, relsep] (isp) at (ilt.base east) {$0.4$};
	\node[decision, fit=(iax)(isp)] (inner) {};
	\node[draw=red!75!black, semithick, ellipse, minimum width=16mm, minimum height=5.4mm, inner sep=0pt] (rule) at (inner.center) {};

	\node[leaf, fill=leafthree] (l3) at (5.0, -1.05) {};
	\node[leafvalue] (l3v) at (l3.center) {$\mu_3$};
	\node[leaf, fill=leafone] (l1) at (0.35, -2.2) {$\mu_1$};
	\node[leaf, fill=leaftwo] (l2) at (1.95, -2.2) {$\mu_2$};

	\draw (root.south) -- node[branch, above left] {YES} (inner.north);
	\draw (root.south) -- node[branch, above right] {NO} (l3);
	\draw (inner.south) -- node[branch, pos=0.4, left] {YES} (l1);
	\draw (inner.south) -- node[branch, pos=0.4, right] {NO} (l2);

	% terminology; each arrow tip touches what it refers to

	\node[ann, anchor=west] (aroot) at (4.2, 0) {Root};
	\draw[annarrow] (aroot) -- (root.east);
	\node[ann, anchor=east] (aax) at (0.1, -0.45) {Splitting axis};
	\draw[annarrow] (aax.east) -- ([shift={(2.5pt, -0.3pt)}]iax.north west);
	\node[ann, anchor=east] (arule) at (-0.25, -1.05) {Decision rule};
	\draw[annarrow] (arule) -- (rule.west);
	\node[ann, anchor=east] (ainner) at (-0.15, -1.8) {Internal node};
	\draw[annarrow] (ainner.east) -- (inner.south west);
	\node[ann, anchor=west] (asp) at (2.35, -1.15) {Splitting point\\(cutpoint)};
	\draw[annarrow] (asp.west) -- ([shift={(-0.6pt, -0.6pt)}]isp.east);
	\node[ann, anchor=west] (aval) at (5.75, -0.6) {Leaf value};
	\draw[annarrow] (aval.west) -- ([shift={(-1pt, -1pt)}]l3v.north east);
	\node[ann, anchor=west] (aleaf) at (5.75, -1.5) {Leaf};
	\draw[annarrow] (aleaf.west) -- (l3.south east);

	% the corresponding partition of the input space

	\begin{scope}[shift={(8.0, -2.55)}, x=2.85cm, y=2.85cm]
		\fill[leafone] (0, 0) rectangle (0.8, 0.4);
		\fill[leaftwo] (0, 0.4) rectangle (0.8, 1);
		\fill[leafthree] (0.8, 0) rectangle (1, 1);
		\draw[region] (0, 0) rectangle (1, 1);
		\draw[region] (0.8, 0) -- (0.8, 1);
		\draw[region] (0, 0.4) -- (0.8, 0.4);
		\node at (0.4, 0.2) {$\mu_1$};
		\node at (0.4, 0.7) {$\mu_2$};
		\node at (0.9, 0.5) {$\mu_3$};
		\node[below] at (0.8, 0) {$0.8$};
		\node[left] at (0, 0.4) {$0.4$};
		\node[below] at (0.4, 0) {$x_1$};
		\node[left] at (0, 0.7) {$x_2$};
	\end{scope}
\end{tikzpicture}
        \caption{\label{fig:tree} Depiction and terminology of a decision tree. Adapted from
                \citet[fig.~1]{hill2020}.}
\end{figure}

A decision tree can be used to represent a parametrized family of functions for
regression. The parameters are the structure of the tree $T$, including decision
rules, and the leaf values $M$. We indicate the function represented by the tree with
$g(\mathbf x;T,M)$. To use this model in Bayesian inference, it is necessary to specify a
prior distribution over $(T,M)$. We factorize the prior as $p(T,M) = p(M\mid T) p(T)$ and
specify the two parts separately.

\paragraph{Prior over tree structure}

The distribution $p(T)$ is defined by the following recursive generative model
(see \cite{chipman1998}, though the grid definition is the one of the \texttt{BayesTree} package that came with \cite{chipman2006,chipman2010}). Fix an axis-aligned grid in the input space, not necessarily
evenly spaced, by specifying a set of splitting points $S_i = \{s_{i1}, \ldots,
	s_{in_i}\}$ for each axis $i \in \{1,\ldots,p\}$. Start generating the tree from the
root, descending as nodes are added. At each new node, consider the subset of splitting
points which are available in its assigned subregion along each axis, and keep only the
axes with at least one split available. If there are none, the node is a leaf. If there
are any, decide with probability $P_d = \alpha/(1 + d)^\beta$ if the node is nonterminal,
i.e., if it should have children, where $d$ is the depth of the node, $d=0$ at the root,
and $\alpha \in [0, 1]$ and $\beta \in [0,\infty)$ are two fixed hyperparameters. If the
node is nonterminal, its splitting axis is drawn uniformly at random from the available
ones, and the splitting point along the chosen axis is drawn uniformly at random from the
available ones.

\paragraph{Prior over leaf values}

If a node is a leaf, its leaf function value is sampled from a Normal distribution with
mean $\mu_\mu$ and standard deviation $\sigma_\mu$, independently of other leaves. This
defines $p(M\mid T)$.

\paragraph{Complete regression model}

The BART regression model uses a sum of $n_\text{tree}$ independent trees, each
with the prior above, plus the usual error term:
\begin{align}
	 &
	y_i = f(\mathbf x_i) + \varepsilon_i
	, \qquad
	f(\mathbf x) = \sum_{j=1}^{n_\text{tree}} g(\mathbf x; T_j, M_j)
	, \label{eq:bartdef} \\
	 &
	\varepsilon_i \mid\sigma \overset{\mathrm{i.i.d.}}\sim \mathcal N(0, \sigma^2)
	, \qquad
	\frac{\lambda\nu}{\sigma^2} \sim \chi^2_\nu
	,                    \\
	 &
	p(\{T_j\}, \{M_j\}) = \prod_{j=1}^{n_\text{tree}} p(M_j \mid T_j) p(T_j).
	\label{eq:bartdef2}
\end{align}
The free hyperparameters in this model are:
\begin{itemize}

	\item The splitting grid. The two common choices are at most 100
		equally spaced splits along each $\mathbf x$ axis in the range
		of the data, or at most 100 splits midway between each observed
		value along each $\mathbf x$ axis.

	\item The number of trees $n_\text{tree}$, common default 200. Although
		indicated with ``$m$'' in the original notation, in this
		manuscript we use $n_\text{tree}$ for clarity.

	\item $\alpha$ and $\beta$, regulating the depth distribution of the
	      trees. Default 0.95 and 2.

	\item $\mu_\mu$ and $\sigma_\mu$, which set the prior mean and variance
		$n_\text{tree}\mu_\mu$ and $n_\text{tree}\sigma_\mu^2$ of
		$f(\mathbf x)$. By default set based on some measure of the
		location and scale of the outcome data.

	\item $\nu$ and $\lambda$, which regulate the prior on the variance of
		the error term. By default $\nu=3$, while $\lambda$ is set to
		some measure of squared scale of the data.

\end{itemize}

\paragraph{The BART MCMC}

Next, we give an overview of the BART MCMC, i.e., the algorithm used to ``learn
the trees''. In this explanation we assume understanding of the basic techniques
used in Bayesian inference; see \autoref{sec:bayesrecap} for a recap of the
topic. For more details see \citet{chipman1998,chipman2010}, \citet[\S
	A]{kapelner2016}, \citet[ch.~5]{daniels2023}, and \citet[\S2.1.4,
	p.~5]{tan2019}.

BART uses Metropolis-Hastings (MH) and conjugate sampling within a Gibbs sampler. The sampler
scans over trees. On each tree, the structure $T_j$ is updated with MH,
conditional on the other trees, but marginalizing the leaves $M_j$ of the
current tree. Then, $M_j$ is conjugately sampled conditional on the other trees
and the updated $T_j$. After scanning all trees, $\sigma^2$ is conjugately
sampled, conditional on the rest. In more detail, each iteration of the MCMC
proceeds as follows. A random small modification is proposed for $T_1$,
proposing to grow a leaf into a node with two leaves, to prune two sibling
leaves making their parent a leaf, or other more complex moves. The change is
either accepted or rejected. Then, $M_1$ is sampled given the potentially
modified $T_1$; this step is a linear regression. This is repeated for all trees
in order. Then $\sigma^2$ is sampled with the trees fixed; its full conditional is
approximately centered on the current residual variance.

Based on our general experience with BART and with developing the implementation
described in this paper, the acceptance of the tree structure Metropolis step is
typically \SI{10}\%--\SI{30}\% \citep[see also][p.~891]{pratola2016}, so each
tree changes on average every 3 to 10 MCMC iterations. Pathological cases
can lower this acceptance to very small values. A rule of thumb on the number of
samples is to run the chain for 2000 samples and discard the first 1000,
differently from the burn-in defaults of popular BART packages, which range from 100 to
500; see also \textcite[6]{kapelner2016}.
%
% default burn-ins on 2024-10-30:
% 100 in BART, dbarts::bart, BayesTree
% 250 in bartMachine
% 500 in dbarts::bart2
%
It is useful to run at least 2 chains and compare the
distributions of the function value at a few points; if the distributions do not
coincide, it may be useful to burn more initial samples. Each tree should not
end up with more than 10 leaves: the algorithm is meant to work with shallow
trees. If the trees grow deeper, increase the number of trees.

This MCMC is quite complex, and moves the tree structures slowly. In practice
in most problems it will not reach convergence, i.e., it won't sample the tree
structures from their actual posterior. However, the function values also
depend on the leaves, which are re-sampled each time from scratch instead of
updated with MH, so the quality of the MCMC samples, while degraded, may be sufficient for
prediction; \textcite[891]{pratola2016} shows an example where, with stuck tree structures, coverage on the latent mean is 54\,\% instead of the 90\,\% target. See \textcite{tan2026computationalefficiencybayesianadditive} for a deeper discussion of convergence.

\section{Implementation of BART for GPU}
\label{sec:impl}

This is the principal contribution of this paper. The implementation described
here corresponds to versions 0.13.0 to 0.14.0 of the Python package \texttt{bartz}
\redactedcite{\citep{gattocrucco_2026_23932737}}. The software is a pure Python
package based on the machine learning library JAX \citep{bradbury2018}, which
provides a computing toolset built around NumPy-like \citep{harris2020} tensor
objects.

Amongst other array-oriented ML frameworks, what makes JAX attractive for
implementing BART is its focus on XLA compilation, useful to write the somewhat
convoluted BART MCMC in simple Python code while avoiding scripting overhead,
and getting good performance both on CPU and GPU. XLA imposes the constraint
that all tensors and their sizes, and often in practice the full sequence of
operations, have to be specified statically, in advance, before running the
program. Thus, our BART implementation in JAX is branchless: the MCMC loop
always executes the exact same set of operations, irrespective of the current
state.

Near-branchlessness is what allows XLA to produce performant programs
automatically. However, the BART MCMC is not a natural fit for branchlessness:
the tree structures change at each iteration, with nodes added and removed, and
the set and number of datapoints assigned to each node altered. To shoehorn
this algorithm into JAX, we make two major deviations from most other
implementations: 1)~we fix a maximum depth for the trees, and then write
tree-traversing operations as if all trees were always maximal, with arithmetic
that masks away operations on nodes that are not actually used; 2)~we always
scan through all datapoints, even for operations that would require only the
subset of datapoints associated with a given node. These sweeping
de-optimizations are feasible because, like boosted trees, the point of BART is
aggregating ``weak learners'', so trees are shallow \citep[see,
e.g.,][\S6]{chipman2010}. In our implementation, we set the maximum tree depth
$D$ to 5 by default, i.e., at most $2^D = 32$ leaves per tree, which is more
than sufficient in normal conditions.

\begin{figure}
	\centering
	\includegraphics[width=\textwidth]{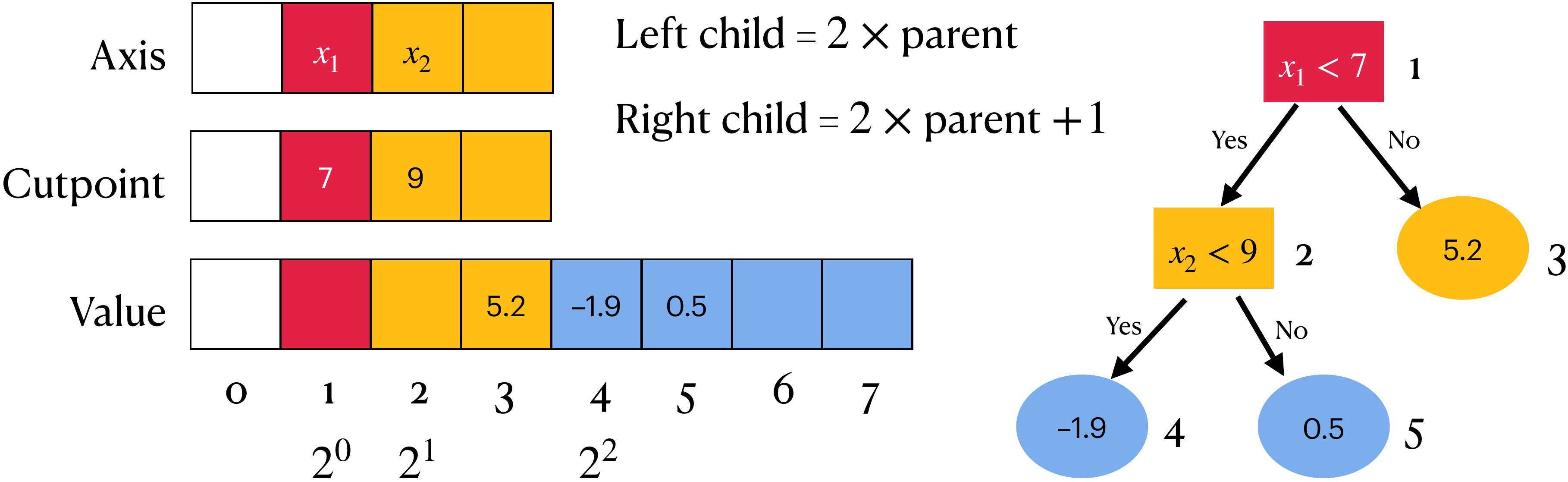}
	\caption{\label{fig:heaptree} Heap representation of a decision tree, with
		the maximum depth set to $D = 2$.}
\end{figure}

To represent the trees as fixed-size matrices, we pad each tree to a perfect tree and list the nodes breadth-first, like a binary heap, as illustrated in
\autoref{fig:heaptree}.
Each tree is represented by 3 arrays, separately holding the axis and splitting
point for decision nodes, and the value for leaf nodes. The arrays are fixed in
size, and each index corresponds to a potential node in the tree. The indices
are assigned to the (potential) tree nodes in order, starting from 1, top to
bottom, left to right (like reading text), imagining a perfect
binary tree with the maximum allowed depth. The entries corresponding to nodes
not actually present in the tree are ignored, as are the axis and cutpoint
entries (resp.\ the leaf value entry) corresponding to leaf (resp.\ decision)
nodes. We fix the convention that the cutpoints start from 1, and that a 0 value
in the cutpoint array entry marks the node as a leaf. The set of trees is then
represented by 3 matrices where each row is a single tree array. See
\autoref{sec:impldetails} for an analysis of the memory usage of this
representation.

Next, we illustrate how to traverse a tree in a branchless fashion, in
\autoref{alg:traverse}. By ``traversing'' a decision tree we mean finding the
\begin{algorithm}[t]
	\KwData{Predictors $\mathbf x$, axis vector $\mathbf a$, cutpoint vector $\mathbf s$}
	\KwResult{Heap index of the leaf $\mathbf x$ falls into}
	leaf\_found = False\;
	index = 1\;
	\For{i = 1 \KwTo $D$}{
		axis = $a_\text{index}$\;
		split = $s_\text{index}$\;
		leaf\_found = leaf\_found OR (split = 0)\;
		child\_index = $2 \times \text{index} + (\text{1 \emph{if} $x_\text{axis} \ge
		\text{split}$ \emph{else} 0})$\;
		index = index \emph{if} leaf\_found \emph{else} child\_index\;
	}
	\Return index\;
	\caption{\label{alg:traverse} Branchless traverse of a decision tree}
\end{algorithm}
leaf associated with a given predictors unit $\mathbf x$. This algorithm is used
after training to compute predictions. To fix in advance the sequence of
operations irrespective of the actual position of the leaf in the tree, we
always loop down to the maximum depth, but we carry a flag to indicate whether
the leaf has already been reached or not, to mask away the redundant operations
on levels below the leaf. JAX automatically vectorizes the code: each operation
in the traverse is repeated for all points and all trees, with scalar variables
becoming vectors or matrices as needed.

In the following we will highlight a few other implementation choices and
important points, without giving a full description of the training algorithm
due to its complexity. For readers familiar with boosting, keep in mind that
this MCMC is kind of like boosting but laying out all trees in advance and then cycling back
in a loop to progressively update the trees in small increments. Suggestively,
it could be called an ``interleaved stochastic tree boosting'' algorithm.

\paragraph{Representation of predictors}

Each predictor $x_i$ can be replaced by its index into the splitting grid. Since
the number of cutpoints rarely exceeds 255 in practice (the default in most BART
packages is 100), each predictor then requires only 1 byte, for a total of $np$
bytes for the matrix of predictors. This quantization is important because the
whole matrix has to fit into GPU memory. This is the same strategy used by,
e.g., LightGBM and CatBoost \autocites{ke_lightgbm_2017}[\S5]{dorogush_catboost_2018}.

\paragraph{Tree structure proposal}

We have to propose a modification to each tree structure as moves for the
Metropolis steps. The original BART picks at random amongst 4 types of moves:
GROW (add two children to a leaf node), PRUNE (remove two sibling leaves), SWAP
(permute the parameters of two directly connected decision nodes), and CHANGE
(change the parameters of a decision node) \citep[\S5.1, p.~940]{chipman1998}.
Following the implementation of the \texttt{BART} package, and against the advice of \textcite[106]{daniels2023}, we use only GROW and
PRUNE for simplicity \citep[\S C, p.~57]{sparapani2021}. To stay branchless, we
have to execute all the calculations for both moves, even if we end up using
only one. However, GROW and PRUNE have the nice properties that one is the inverse of the other, the proposed moves can not
overlap in the tree arrays, and the tree proposed by GROW contains the
initial tree which in turn contains the tree proposed by PRUNE. This allows
sharing memory and most of the calculations between the two moves, and
simplifies the updates to the indices cache (described below) and the running
residuals.

\paragraph{Leaf indices cache}

The indices that, for each tree, indicate which leaf each datapoint falls
into, are used to compute the acceptance probability of the Metropolis steps,
and to sample the leaves. At each MCMC iteration, these indices change in
limited ways due to the simplicity of the GROW and PRUNE moves, so it is
convenient to keep the indices in memory rather than recompute them from scratch
each time with \autoref{alg:traverse}. If the maximum depth of the trees $D$ is
$\le 7$, the indices fit into a byte, so this cache uses $n \times
	n_\text{tree}$ bytes (per chain). Together with the $n\times p$ predictors matrix, this is
the bulk of the memory usage of this algorithm, and it is an
additional memory requirement compared to (non-Bayesian) boosting. We estimate
removing this cache would slow training down by 2--3x.

\paragraph{Parallelization along trees}

The key steps of the MCMC that are computed in parallel across trees are:
\begin{enumerate}
	\item Propose tree structure moves.
	\item Update the leaf indices cache.
	\item Sample centered full-conditional leaves.
	\item Compute most of the terms in the Metropolis acceptance probabilities.
\end{enumerate}

Instead, the computations that have to follow the ordered tree scan are:
\begin{enumerate}[start=5]
	\item Sum the partial (i.e., all trees less the current one) residuals
	      falling in each leaf (an indexed reduce); this is the main bottleneck at large $n$.
	\item Complete the calculation of the Metropolis acceptance probability and
	      accept/reject the move.
	\item Compute the posterior mean of the leaves to un-center the sampled leaf values.
	\item Update the running residuals.
\end{enumerate}

These steps can not be run in parallel for all trees due to the cyclical
dependence between running residuals and the acceptance status of each move.
Since this part includes two $O(n)$ operations (residuals reduction and update),
it can be parallelized effectively only if $n$ is large enough. As shown in
\autoref{sec:perf} next, this can be a problem in practice, though
we mainly care about accelerating BART on large datasets.

\paragraph{Additional features}

In addition to the vanilla BART model and MCMC, \texttt{bartz} v0.14.0
implements classification with probit link \autocite[\S4]{chipman2010}, variable selection
\citep{linero2018}, data parallelism
\autocite[shard datapoints across devices,][]{pratola2014}, chain parallelism (run independent MCMC chains across
devices), multivariate outcomes, heteroskedasticity with fixed weights, and Bayesian Causal Forests \autocite[BCF,][]{hahn2020}.

\section{Performance measurements}
\label{sec:perf}

We measure the running time and the prediction accuracy of our BART
implementation \texttt{bartz} together with other BART packages and
gradient boosting packages, on real and simulated data.

\paragraph{Data generating process}

We run some of the benchmarks on synthetic data to conveniently set training
set size $n$ and number of predictors $p$ to arbitrary values while also
controlling the signal-to-noise ratio (SNR). The data generating process (DGP) is the
following sparse quadratic model:
\begin{align}
	X_{ij}                               & \overset{\mathrm{i.i.d.}}\sim U(-\sqrt 3, \sqrt 3), \quad i = 1, \ldots, n, \quad j = 1, \ldots, p,               \\
	\beta_j, A_{jj'}                     & \overset{\mathrm{i.i.d.}}\sim \operatorname{Rademacher}, \\
	s_j & \overset{\mathrm{i.i.d.}}\sim \operatorname{Bernoulli}(p_\text{eff}/p), \quad p_\text{eff}\in(0, p], \\
	P_{jj'}                              & = \begin{cases}
		                                         1 & \min(|j-j'|,p-|j-j'|) \le q/2, \\
		                                         0 & \text{otherwise,}
	                                         \end{cases} \quad q \bmod 2 = 0, \quad q < p,                                                                  \\
	\mu(x) &= \sum_j
	\frac{\sigma_\text{lin}}{\sqrt{p_\text{eff}}}
	s_j \beta_j x_j + \sum_{jj'}
	\frac{\sigma_\text{quad}}{\sqrt{p_\text{eff}(4/5 + q\,p_\text{eff}/p )}}
	s_j s_{j'} P_{jj'} A_{jj'} x_j x_{j'}, \\
	Y_i\mid X_{i\cdot},s,\beta,A           & \sim \mathcal N\left(\mu(X_{i\cdot}), \sigma^2_\text{eps}\right). \label{eq:dgp}
\end{align}
The model definition is chosen to yield these convenient moment relationships:
\begin{align}
	E[\operatorname{Var}[Y_i|s,\beta,A]] & = \sigma^2_\text{lin} + \sigma^2_\text{quad} + \sigma^2_\text{eps}, \quad \text{(expected population variance)} \label{eq:popvar}  \\
	\operatorname{Var}[Y_i] & = E[\operatorname{Var}[Y_i|s,\beta,A]] + \frac{\sigma^2_\text{quad}} {4/5 + q\,p_\text{eff}/p }, \quad \text{(``prior'' variance)} \label{eq:priorvar}
\end{align}
and we set the variance hyperparameters as follows, yielding $\text{SNR}=2$ on average:
\begin{align}
	\sigma^2_\text{lin} &= \sigma^2_\text{quad} = \sigma^2_\text{eps} = 1/3.
\end{align}

\paragraph{Matrix of dataset and model sizes}

The running time and memory usage are mostly determined by $n$, $p$ and the
number of trees, so we repeat most benchmarks on a power-of-2 scan of $n$ by a
$2\times 2$ grid of settings: low/high $p$ $\times$ low/high number of trees. As
``low'' settings we use $p=100$ or $p=50$, $n_\text{tree} = 200$, while as
``high'' we make the parameters grow proportionally to $n$ with $n/p = 10$ or
$20$ and $n/n_\text{tree}=8$. The number of trees being proportional to $n$ is
an unusually high setting. These combinations are not optimal for inference,
they are simple extreme cases to explore the computational complexity of the
algorithms. The ``low-low'' setting is the one closer to common usage. In this
matrix, we use the DGP hyperparameters $p_\text{eff}=p$ (all predictors active)
and $q=2$ (each predictor interacts with one other predictor on average).

\paragraph{Single-iteration timing measurements}

\begin{figure}
	\centering
	\widecenter{\includempl{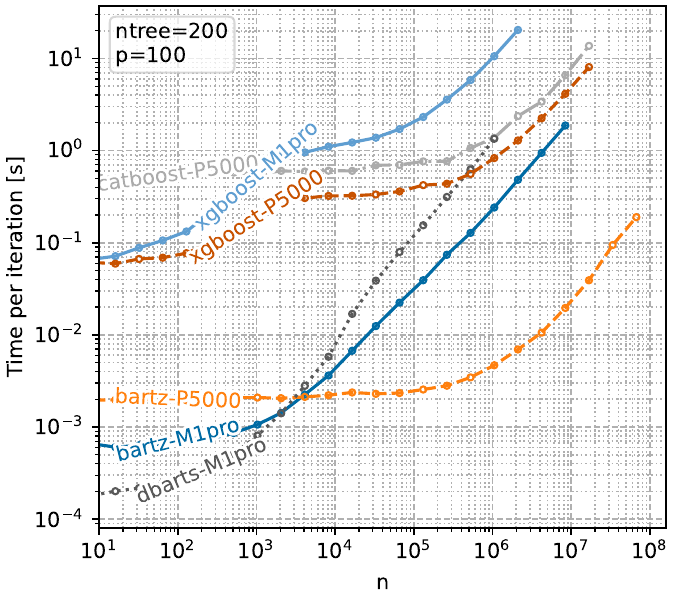}\includempl{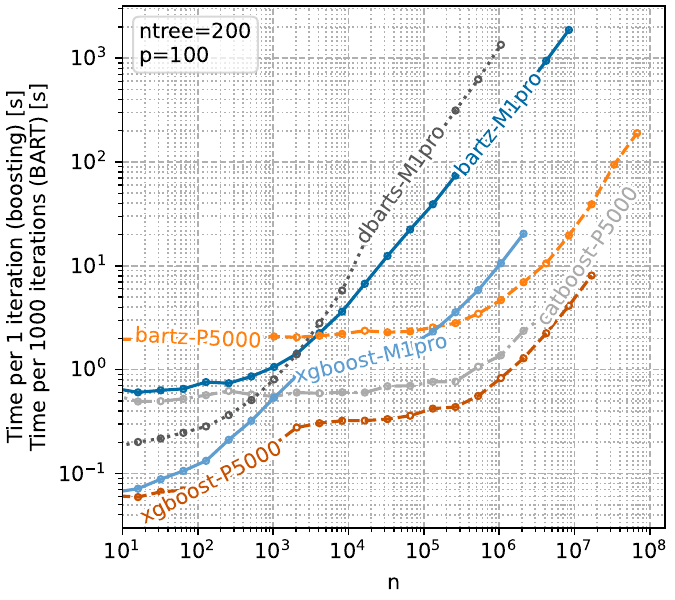}}
	\caption{\label{fig:single} \textbf{Left panel:} Time to run an
		iteration (one sweep through all trees) of the algorithm vs.\ training set size $n$,
		comparing \texttt{bartz} with the previous fastest BART implementation
		(\texttt{dbarts}), XGBoost, and CatBoost, on
		CPU (\texttt{-M1pro}) and GPU (\texttt{-P5000}). Gradient
		boosting requires only one iteration to produce a usable
		result, while BART requires $O(1000)$ iterations, so the
		\textbf{right panel} shows the same data but with BART timings
		multiplied by an arbitrary factor 1000 to aid visual comparison
	of the lines. See Figures~\ref{fig:time-all}
and~\ref{fig:time-all-1000} for the complete results.}
\end{figure}

We pit our implementation \texttt{bartz} on CPU and GPU against \texttt{dbarts}
(the previous fastest BART implementation), XGBoost, and
CatBoost on data generated as \autoref{eq:dgp}. Our CPU is an Apple M1
Pro with 16 GiB RAM, while the GPU is an Nvidia RTX PRO 5000 Blackwell with 48
GiB RAM. All methods but \texttt{dbarts} are allowed to use up to 6 cores on CPU.
For all software packages, we run their setup and pre-processing step
separately, before starting the timer. This includes data pre-processing, and
XLA compilation for \texttt{bartz} specifically. The configurations are mostly
left to their default values. We clock a full iteration (a sweep through all
trees), taking the best of three runs. For BART packages, we repeat the run
starting from the final state of the previous run. For gradient boosting
packages, each run starts fresh. While \texttt{bartz} is branchless,
\texttt{dbarts} is not, and given the state dependence, \texttt{dbarts} timing
is less reliable. Since we take best of three, the bias should be in the
conservative direction of making \texttt{dbarts} appear faster.
See \autoref{sec:timingdetails} for further details on the setup.

The results with the ``low-low'' settings $p=100$, $n_\text{tree}=200$ are
shown in \autoref{fig:single}. On CPU, \texttt{dbarts} is 2--3
times faster than \texttt{bartz} at low training set size $n$, then
\texttt{bartz} takes over in the range $n=\num{1000}\text{--} \num{10000}$ and
reaches a 5x advantage at large $n$. On GPU, \texttt{bartz} switches from a
constant to a linear regime around $n = \num{1000000}$, highlighting the
tree-serial bottleneck that requires parallelization along $n$ discussed in
\autoref{sec:impl}. At high $n$, \texttt{bartz} on GPU is 100x faster than
itself on CPU, by extrapolation 500x faster than the previous state of the art
\texttt{dbarts}, and 200x faster than XGBoost on GPU. Notably, XGBoost is only
20x faster than itself on GPU vs.{} CPU, and our CPU runtime beats the previous
state of the art by 5x, so the comparison is not artificially inflated by a
slow CPU version. Considering a BART run with 2000 MCMC iterations, it would
then be 10x slower than a single XGBoost run GPU vs.{} GPU (at same number of
trees), while on CPU vs.{} CPU it would be 50x slower. (Note again: these
figures do not include predictor quantization.)

BART requires 1000--2000 (traditionally 1100) iterations to produce a result,
while gradient boosting only one iteration, so comparing per-iteration timings
across the methods requires multiplying by some factor. However this factor is
not precisely defined because it's common to cross-validate gradient boosting
methods, it's also common to run BART longer to be sure the MCMC is converging,
and the optimal number of trees may differ between the methods. So in
\autoref{fig:single} we show side-by-side the actual iteration time, and
iteration time scaled by 1000 on BART methods, where 1000 is an arbitrary
factor chosen to align well with the logarithmic scale of the plot. Another
reason it's hard to interpret these timings precisely is that the running time
of XGBoost and CatBoost (with default settings) scales with $p$, while BART
doesn't (the search is implicitly amortized in the MCMC iterations).

\paragraph{Predictive accuracy comparison with other BART packages}

\begin{figure}
	\centering
	\widecenter{\includempl{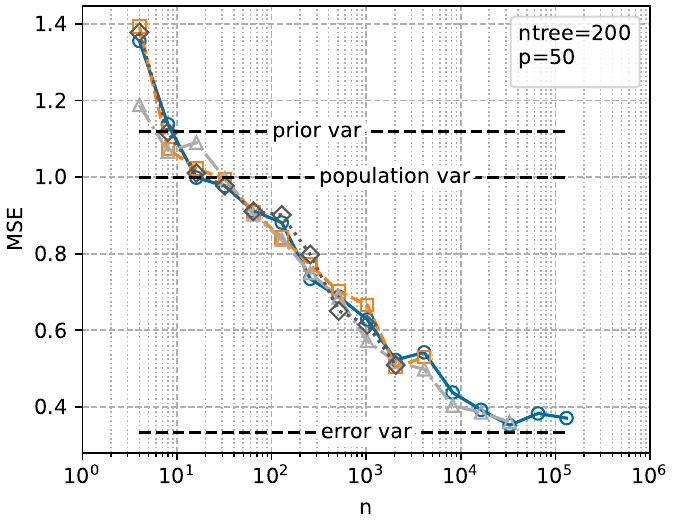}\includempl{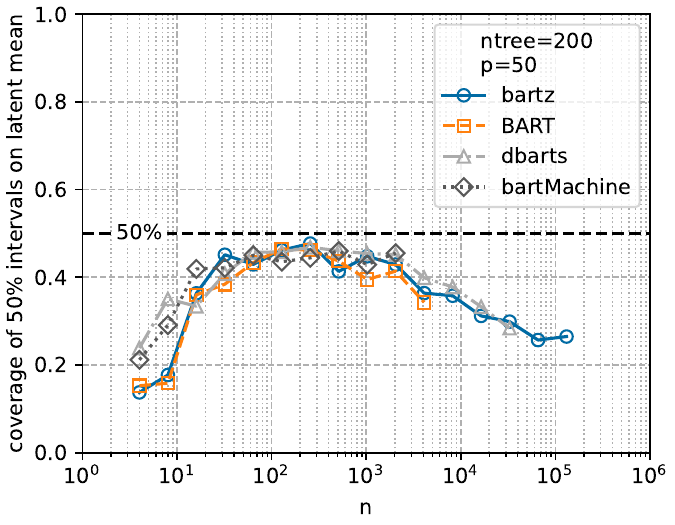}}
	\caption{\label{fig:accuracy-single} Test set mean squared error (MSE)
		(\textbf{left panel}) and coverage of \SI{50}\% intervals on
		the latent mean (\textbf{right panel}) vs.\ training set size,
		comparing \texttt{bartz} with the most popular implementations
		of BART, to check they yield similar results. See
	Figures~\ref{fig:rmse-all} and~\ref{fig:coverage-all} for the full
results.}
\end{figure}

Since we re-implemented the well-known BART MCMC without substantial changes,
we do not include a systematic assessment of the predictive performance of BART itself
\autocites[see for example][table~2]{linero2018b}[supplementary]{he2019}[table~4]{he_stochastic_2023}[fig.~5]{nguyen2025oblique}. To show at least some
evidence here that our implementation is correct, we compare the out-of-sample
prediction accuracy between our package \texttt{bartz} and the 3 most popular
BART implementations, on data generated as \autoref{eq:dgp}. The result at
``low-low'' settings is in \autoref{fig:accuracy-single} (complete results in
\autoref{sec:compdetails}); mean squared error and coverage shake out
similar across packages as expected.
Note: in \autoref{fig:accuracy-single} we show coverage of credible intervals
for prediction of the latent mean rather than intervals for prediction of the
observable outcome, because targeting a confidence level on exchangeable data
is too easy and so it would not have discriminative power.

\paragraph{Full run comparisons}

\begin{table}
	\centering
	\caption{\label{tab:fullbench} Comparison of \texttt{bartz} against
		XGBoost on various large datasets,
		on GPU. ``xgboost-es'' uses early stopping with a
		large maximum number of trees, and ``bartz-2000'' uses 2000
		trees instead of the default 200.
		Each run is repeated 5 times with different random seeds,
		the standard deviation across runs is reported in parentheses. Time in seconds.
		``life'' is the number of iterations since each tree was last
		pruned back to a root, averaged over trees at the end of the
	MCMC, and divided by the total number of MCMC iterations. See
\autoref{tab:fullbench-full} in the appendix for the complete results.}
	\small\sffamily
	% typeset the numbers as text, in the current font family
	\sisetup{mode=text, reset-text-family=false}
	\setlength{\tabcolsep}{3pt}
	\widecenter{\begin{tabular}{lrrlrrrrrrrrrr}
\toprule
 &  &  &  & \multicolumn{2}{c}{xgboost} & \multicolumn{3}{c}{bartz} & \multicolumn{2}{c}{xgboost-es} & \multicolumn{3}{c}{bartz-2000} \\
\cmidrule(lr){5-6} \cmidrule(lr){7-9} \cmidrule(lr){10-11} \cmidrule(lr){12-14}
dataset & \textit{n} & \textit{p} & task & RMSE & time & RMSE & time & life & RMSE & time & RMSE & time & life \\
\midrule
electricity & \num{1145264} & \num{11} & regr. & \num{0.136(2)} & \num{0.4} & \num{0.138(2)} & \num{19} & \num{1.00} & \num{0.119(2)} & \num{2.7} & \num{0.131(2)} & \num{96} & \num{0.80} \\
higgs & \num{10990000} & \num{28} & class. & \num{0.414(3)} & \num{1.6} & \num{0.417(2)} & \num{74} & \num{1.00} & \num{0.397(2)} & \num{37.3} & \num{0.403(2)} & \num{537} & \num{0.95} \\
radar traffic & \num{1958768} & \num{14} & regr. & \num{0.258(2)} & \num{0.5} & \num{0.272(1)} & \num{25} & \num{1.00} & \num{0.208(5)} & \num{8.6} & \num{0.227(3)} & \num{135} & \num{0.96} \\
zurich delays & \num{5455575} & \num{20} & regr. & \num{0.946(7)} & \num{0.7} & \num{0.946(7)} & \num{60} & \num{0.99} & \num{0.942(6)} & \num{7.8} & \num{0.945(7)} & \num{333} & \num{0.57} \\
poker & \num{1012616} & \num{5} & class. & \num{0.32(1)} & \num{0.3} & \num{0.116(6)} & \num{18} & \num{1.00} & \num{0.044(5)} & \num{1.9} & \num{0.076(3)} & \num{97} & \num{0.92} \\
simulated & \num{3990000} & \num{1000} & regr. & \num{0.707(3)} & \num{11.1} & \num{0.790(4)} & \num{43} & \num{1.00} & \num{0.621(2)} & \num{51.3} & \num{0.601(3)} & \num{223} & \num{0.76} \\
\bottomrule
\end{tabular}
}
\end{table}

Finally, having shown that our BART implementation is the fastest on large datasets,
we compare \texttt{bartz} against XGBoost on GPU
on full runs on a few large datasets. We select 5 large datasets from OpenML \autocite{OpenML2025},
and simulate one large dataset from \autoref{eq:dgp} with $p_\text{eff}/p =
0.1$ and $q=32$. We run XGBoost and \texttt{bartz} with the default settings of their
high-level interfaces (no cross-validation), but for using a single MCMC chain in \texttt{bartz} instead of
the default four, and optimizing XGBoost data loading for GPU. See \autoref{sec:fulldetails} for more details.
 
The results are in \autoref{tab:fullbench}.
On the largest dataset (``higgs'',
$n\approx\num{11000000}$), \texttt{bartz} is 46x slower than XGBoost, roughly in line
with the large-$n$ 10x factor previously estimated from \autoref{fig:single}:
it's 10x at same number of trees, which becomes 20x with XGBoost's default 100 trees, then 70x going from $p=100$ to $p=28$.
On smaller datasets \texttt{bartz} is 50--80x slower, except for the simulated
dataset, where \texttt{bartz} is only 4x slower than XGBoost. What sets our simulated dataset
apart from the others is that it has many predictors, $p=1000$ vs.{}
$p=\numrange{5}{28}$. As expected, BART is advantaged in high-$p$ cases, at
least with the default XGBoost settings.

The BART literature claims that, without tuning, BART is better than XGBoost,
while here it's overall a draw at best: \texttt{bartz} has nearly the same RMSE as XGBoost on 3
datasets (``electricity'', ``higgs'', ``zurich delays''), worse RMSE on two
(``radar traffic'', ``simulated'') and better RMSE on one (``poker'').
Interestingly, ``poker'' is the only
almost deterministic dataset in the bunch. The difference between the
pre-existing benchmarks and ours is size: this is the first time BART is
systematically benchmarked on datasets with $n$ in the millions.

In this context, the main hyperparameter of both \texttt{bartz} and XGBoost is the
number of trees. The defaults, respectively 200 and 100 trees, are a priori not
enough on datasets this large, since both algorithms are designed to work with
shallow trees (2--10 leaves per tree in BART) and the total number of leaves
must increase with the complexity and resolution of the signal to explain. So
we also let XGBoost choose the optimal number of trees, and run \texttt{bartz} with 2000
trees (``xgboost-es'' and ``bartz-2000'' in \autoref{tab:fullbench}). We also
show ``life'', a measure of how long-lived the trees are in the BART MCMC, to
check whether the MCMC is mixing well.

Overall, ``bartz-2000'' appears worse than ``xgboost-es'': ``xgboost-es'' has
RMSE better than ``bartz-2000'' on the 5 OpenML datasets, and worse
on the simulated high-$p$ one. The 200-trees \texttt{bartz} run has average tree life at
or close to the maximum 1.0, which means the trees are almost never completely
reset during the course of the MCMC. The 2000-trees run instead has tree lives
in 0.57--0.96, better but still quite high, suggesting the optimal number of
trees for BART may be even higher on these datasets. To fit our compute budget,
we didn't optimize the number of \texttt{bartz} trees via cross-validation, and BART has
no practical equivalent to XGBoost's early stopping (the technique in
\cite[ch.~3]{collins_improved_2023} would
require further research to scale to large datasets).

\section{Conclusions}
\label{sec:conclusions}

We provide the first GPU implementation of BART as a Python package,
\texttt{bartz}. On large datasets, we surpass the previous state of the art
(\texttt{dbarts}) on CPU by 5x, and achieve a further 100x speed-up on GPU (with
some given CPU and GPU models), for a total 500x. This makes 2000 iterations of
the BART MCMC about 10x slower than one XGBoost run (excl.\ predictor
quantization, same number of trees, at $p=100$), while the analogous factor on CPU would be
50x with \texttt{bartz} and would have been 250x with \texttt{dbarts}. Beside
the absolute speed-up, this suggests that BART might in general parallelize better
than non-Bayesian tree learning. Our package beats the previous state of the
art on CPU and is written in pure Python without custom GPU kernels, so this
observation is not simply an artifact of comparing a good GPU implementation to
a poor CPU one.

XBART, the previous most successful attempt at accelerating BART,
achieved at most a 20--30x speed-up at the time \autocite[table~4, p.~560]{he_stochastic_2023}, but at the cost
of changing the algorithm and not being an implementation of BART but rather a
BART-inspired non-Bayesian algorithm. Note again: with ``previous state of the
art'' above we refer to the BART implementation \texttt{dbarts} rather than
XBART.

Our speed-up significantly increases the scope of application of BART. While our
ambition would ultimately be to make Bayesian methods as common as non-Bayesian
methods in this class, we note that raw speed is not sufficient:
popular gradient boosting libraries are
feature-packed high-quality software packages, not matched by the academic
implementations of BART, including ours. Furthermore, BART has no fast
method to infer the number of trees. Because of this, we were not able to show
a general BART advantage on RMSE on large ($n$ in the millions) datasets. It's
plausible the known BART performance on small datasets would generalize to
large ones, but it currently stands as an open question, and suggests the most
important line of research is to make inference on the number of trees
scalable, possibly starting from \textcite{collins_improved_2023}.

We wish to also highlight some limitations of BART on GPU in absolute terms
instead of relative to non-Bayesian methods. BART is quite good at dealing with
high-$p$ data and variable selection, so it is a natural question how much this
GPU implementation could push BART up along $p$ (number of predictors) rather
than $n$ (number of datapoints). Alas, our implementation runs out of GPU
memory when the number of predictors $p$ or the number of trees are high enough
(see \autoref{sec:gpumem}). While the limitation on the number of trees could
be avoided with a compromise on speed by removing the leaf indices cache, the algorithm
requires random access to the predictors, so all of them must reside in memory.
For the same reason, data parallelism is worthwhile only sharding $n$, not $p$,
thus, if multiple GPUs were employed to keep a very-high-$p$ predictors matrix
in memory, each GPU would end up with a small $n$, blocking effective
parallelization.

This indicates that, beside optimization at high $n$, low-$n$ speed-ups are
also valuable. A potential solution could be to implement XBART on GPU: since
it scans through predictors or subsets of predictors, instead of accessing them
at random, we expect it would allow complementing parallelization/sharding
along $n$ with parallelization/sharding along $p$. Another potential solution
could be to devise variations of the BART model and MCMC that do not require
fully random access to all predictors in all trees. A conceptually simpler
direction could be to further optimize our implementation for low-$n$ speed,
but there is probably an intrinsic limit on how small $n$ can be.

\begin{ack}

	We thank Antonio Linero, Andrew Herren, P.\ Richard Hahn, and Francesco
	C.\ Stingo for useful comments and discussions. This work was done at
	the Department of Statistics, Computer Science, Applications (DISIA)
	``G.~Parenti'' of the University of Florence (UNIFI), Italy, and at the
	Department of Statistics and Data Science (SDS) of the University of
	Texas at Austin (UT Austin), USA. The introductory material in
	sections~\ref{sec:intro} and~\ref{sec:bart} is adapted from
	\redactedcite{\citet{petrillo2024f}} by the same author. We thank
	Miaoqing Yu for implementing additional features in the software
	package. Various Claude models greatly helped improve, revise and
	polish this work, but the prose and original software implementation
	are human-written. The author received no specific funding for this
	work, and declares no competing interests.

\end{ack}

\printbibliography

%%%%%%%%%%%%%%%%%%%%%%%%%%%%%%%%%%%%%%%%%%%%%%%%%%%%%%%%%%%%%%%%%%%%%%%%%%%%%%%
%%%%%%%%%%%%%%%%%%%%%%%%%%%%%%%%%%%%%%%%%%%%%%%%%%%%%%%%%%%%%%%%%%%%%%%%%%%%%%%
% APPENDIX
%%%%%%%%%%%%%%%%%%%%%%%%%%%%%%%%%%%%%%%%%%%%%%%%%%%%%%%%%%%%%%%%%%%%%%%%%%%%%%%
%%%%%%%%%%%%%%%%%%%%%%%%%%%%%%%%%%%%%%%%%%%%%%%%%%%%%%%%%%%%%%%%%%%%%%%%%%%%%%%
\clearpage
\appendix

\section{Recap on Bayesian inference with MCMC}
\label{sec:bayesrecap}

In the Bayesian paradigm, the parameters of the model (and consequently the
predictions) are inferred from the data by conditioning a joint data-parameters probability
distribution on the observed values of the data, obtaining a posterior
distribution for the parameters. The BART model definition in
Equations~\ref{eq:bartdef}--\ref{eq:bartdef2} can be expanded into a joint
probability density function of the data and parameters $p(\mathbf y, \sigma^2,
	\{T_j\}, \{M_j\})$, $\mathbf y = (y_1, \ldots, y_n)$; the posterior $p(\sigma^2, \{T_j\}, \{M_j\}\mid \mathbf y)$
is, up to a constant multiplicative factor, the same expression but with
$\mathbf y$ held fixed at the observed values. This distribution expresses the
degree of belief in each possible combination of parameter values being the
correct one; e.g., the mode or the mean of the distribution could be used as
estimates of continuous parameters.

\paragraph{MCMC}

Even though the posterior $p(\sigma^2, \{T_j\}, \{M_j\}\mid \mathbf y)$ could be
written down explicitly, it is a long and complex formula that is in practice
impossible to interpret. To represent the posterior in a legible way, the most
common technique is Markov chain Monte Carlo (MCMC): setting $\theta=(\sigma^2,
	\{T_j\}, \{M_j\})$ for brevity, it is possible to define an iterative numerical
procedure that, starting from an initial choice of parameter values
$\theta^{(0)}$, evaluating numerically the joint density function
$\theta\mapsto p(\mathbf y,\theta)$ only at some points, picks a new set of
values $\theta^{(1)}$, and so on $\theta^{(2)}, \theta^{(3)}, \ldots$, yielding
overall a sequence of parameter values which is a random sample from the
posterior. Then, e.g., the mean of the distribution is approximated by the
average of these samples. See \citet{brooks2011} for a reference on MCMC.

The samples produced by MCMC are only asymptotically correct: the chain takes
some time to ``forget'' the arbitrary initial state $\theta^{(0)}$ and start
sampling from the target distribution $p(\theta|\mathbf y)$. They are also not
independent, as each sample is picked starting from the previous one. In
practice, some of the initial samples must be discarded, and even after that,
the remaining number of MCMC samples must be higher by some factor than what's
sought.

\paragraph{Metropolis-Hastings and Gibbs sampling}

A common general way of building an MCMC is the Metropolis-Hastings (MH) class of
samplers. An MH sampler requires defining a \emph{proposal distribution}
$p(\theta';\theta)$. At MCMC step $k$, the current values of the parameters
$\theta^{(k)}$ in the chain are plugged as parameters of the proposal as
$p(\cdot;\theta^{(k)})$. A random draw $\tilde\theta^{(k+1)}\sim
	p(\cdot;\theta^{(k)})$ from the proposal gives potential new parameter values
$\tilde\theta^{(k+1)}$. Then a criterion decides if the new proposed values are
accepted (keep new values, $\theta^{(k+1)} = \tilde\theta^{(k+1)}$) or rejected
(keep current values, $\theta^{(k+1)} = \theta^{(k)}$), with a biased coin flip
that depends on the posterior and favors the choice of $\theta^{(k+1)}$ that
makes the posterior density higher, but adjusting for how much the proposal is biased towards going there. In other words, the sampler tries to move
one step at a time in parameter space, picking a move at random, testing the
ground, and deciding whether to proceed in that direction, based on how much the
probability changes going there.

A special case of Metropolis-Hastings is Gibbs sampling. Split the parameters into
two subsets $\theta = (\theta_a, \theta_b)$. If we adopt an MH proposal that
proposes changes to $\theta_a$ only, keeping $\theta_b$ fixed, from the
\emph{full conditional} distribution $p(\theta_a|\theta_b,\mathbf y)$, it turns
out that the MH acceptance probability is always~1 and the move is always
accepted. Alternating this procedure between $\theta_a$ and $\theta_b$ yields a
valid sampler for the full $\theta$. Gibbs sampling allows nesting sampling
strategies, by applying them one parameter at a time with the full conditional
as posterior distribution.

With these notions in hand, the high-level description of an iteration of the BART MCMC is
\begin{align}
	\text{loop on tree $j$: } & \begin{cases}
		                            T_j \sim \text{MH on } p(T_j\mid\mathbf T_{-j},\mathbf M_{-j},\mathbf y,\sigma^2), \\
		                            M_j \sim p(M_j\mid\mathbf T,\mathbf M_{-j},\mathbf y,\sigma^2),
	                            \end{cases} \\
	\sigma^2                  & \sim p(\sigma^2\mid\mathbf T,\mathbf M,\mathbf y),
\end{align}
where the subscript ``$-j$'' means ``all indices but $j$''.

\section{Further details on the implementation}
\label{sec:impldetails}

\paragraph{Size of the tree representation}

With reference to \autoref{sec:impl}: when trees are not full, the fixed-size heap representation of the trees
wastes most of the array entries. However it turns out to use about the same
memory as other common alternatives. The maximum number of nodes is $2^{D+1} - 1$,
so the heap size is $2^{D+1}$ (index 0 is unused). Since the nodes at maximum depth
can only be leaves, the axis and cutpoint arrays are only $2^D$ long. Using
8 bit integers for the cutpoints, 16 bit integers for the axes, and 16 bit floats for the leaves (these are the sizes used by \texttt{bartz} on GPU if $p > 256$), the total size in bytes is $2^D +
	2\cdot2^D + 2\cdot 2^{D+1} = 7\cdot 2^D$ bytes. If $D=5$, that's 224 bytes per tree. Consider instead
a representation in C/C++ that used separately allocated linked node objects.
Each node contains allocation overhead (8 bytes), axis (2), cutpoint (1), leaf
value (2), and pointers to children ($2\times 8$), for a total of 29 bytes per
node, rounded to 32 for alignment. This matches the size of the heap if there
are 7 nodes per tree on average, and it's also not contiguous in memory. Other
possible linked schemes, and those used in existing implementations, similarly
consume a lot of memory in scaffolding; for example, \texttt{BART} uses 64 bytes
per node. % see BART 2.9.6, files src/tree.h:100, src/tree.cpp:56

\paragraph{Numerical error due to caching residuals}

Since the residuals are cached and updated according to tree modifications,
instead of recomputed, a potential problem is the accumulation of numerical
error. The values of the residuals and their updates are similar across
datapoints and trees, so, with 32 bit floats, each MCMC iteration adds a
relative numerical error with standard deviation $\approx \num{1e-7} \sqrt
	{n_\text{tree}}$. This is not a problem for any current BART application, but
it would be with 16 bit floats, which might be an option to further accelerate
the implementation.

\section{Details of single-iteration timing measurement}
\label{sec:timingdetails}

\begin{figure}
	\widecenter{\includempl{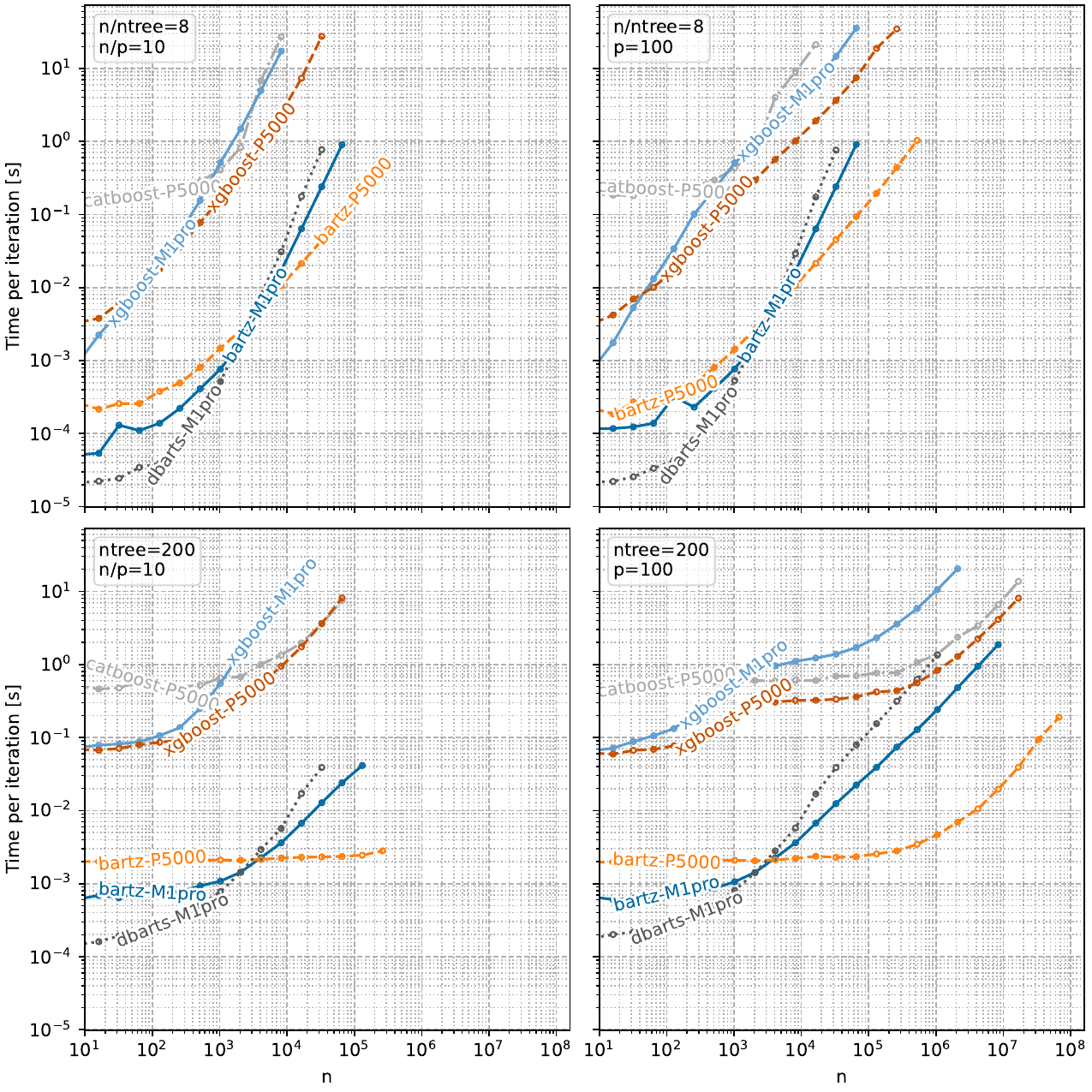}}
	\caption{\label{fig:time-all} Full version of \autoref{fig:single},
		left panel. Time to run an iteration (one sweep through all
		trees) of the algorithm vs.\ training set size $n$, comparing
		\texttt{bartz} with the previous fastest BART implementation
		(\texttt{dbarts}), XGBoost, and CatBoost, for
		various settings of number of trees and number of
		predictors~$p$, on CPU (\texttt{-M1pro}) and GPU
		(\texttt{-P5000}). Take into account that gradient boosting
		requires only one iteration to produce a usable result, while
		BART requires $O(1000)$ iterations. See
		\autoref{fig:time-all-1000} for a version of this plot where
		the BART running time is multiplied by 1000.}
\end{figure}

\begin{figure}
	\widecenter{\includempl{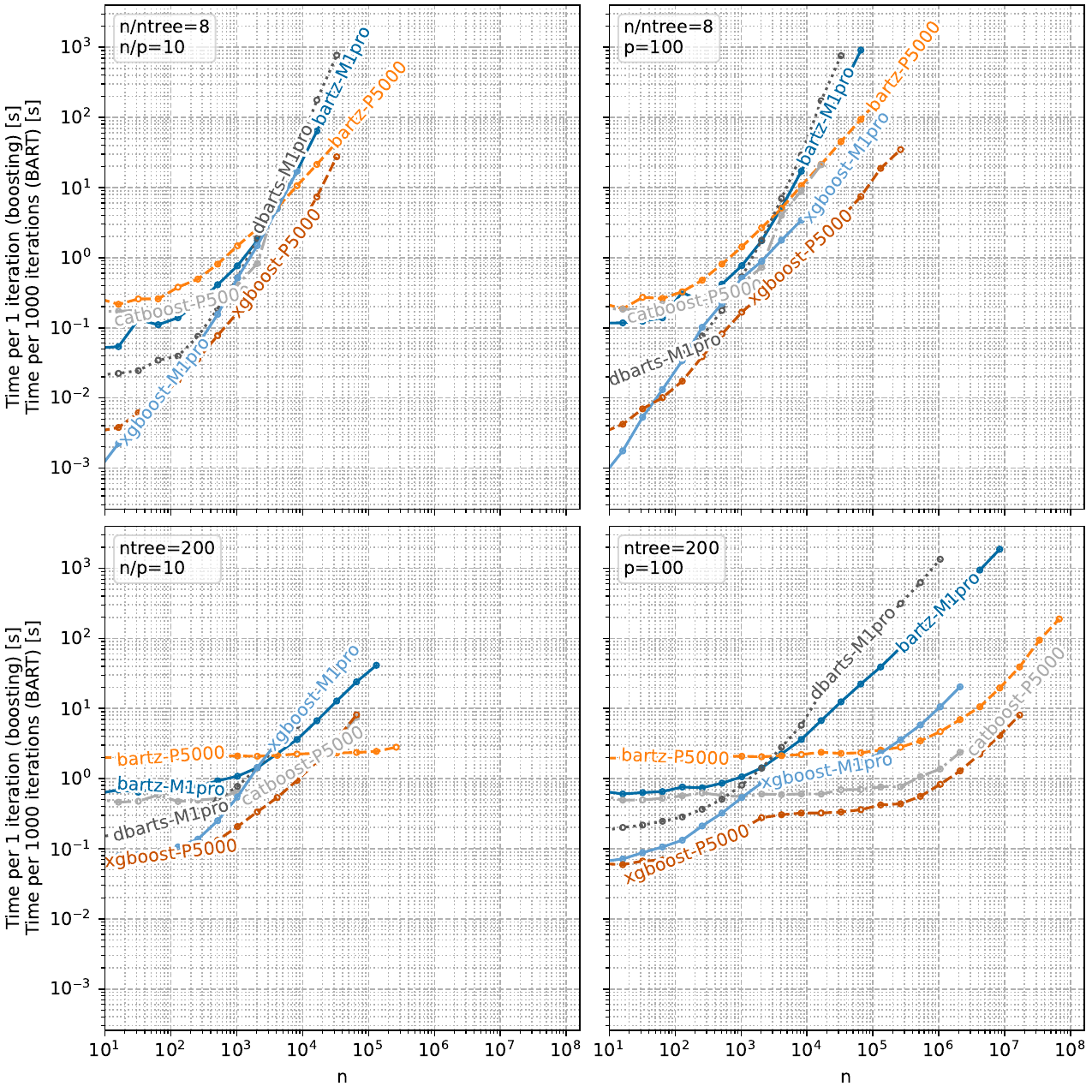}}
	\caption{\label{fig:time-all-1000} Full version of
		\autoref{fig:single}, right panel. Time to run 1 iteration  of
		XGBoost, CatBoost, or 1000 iterations of
		\texttt{bartz}, \texttt{dbarts}, vs.\ training set size $n$.
		This is the same data as \autoref{fig:time-all}.}
\end{figure}

The single-iteration timing benchmarks for all combinations of
$p$-$n_\text{tree}$ settings are in Figures~\ref{fig:time-all}
and~\ref{fig:time-all-1000}. Further details on the setup:

\begin{itemize}

	\item BART packages (\texttt{dbarts} and our \texttt{bartz}) are timed
		by letting their MCMC scaffold run 15 MCMC iterations. This is
		repeated 3 times, and the time of the fastest round is divided
		by 15 to get the quoted measurement.

	\item XGBoost and CatBoost are timed by repeating three times a single
		run (building all trees from scratch), then taking the fastest
		of the three rounds.
	
	\item BART packages start each successive round from the MCMC state of
		the previous one, while boosting packages always start from
		scratch. The initial MCMC state is the default one, with empty
		trees.

	\item For all packages but CatBoost, we only time the part that's
		repeated over trees, instead of the pre- and post-processing
		that does not depend on the number of trees. The CatBoost API
		does not allow us to do this completely, though predictor
		quantization remains separate, which is probably most of the
		fixed overhead.

	\item \texttt{dbarts} has an \texttt{n.threads} parameter to use
		parallelization, but we found we could get at most a 1.3x
		speed-up at $n \ge \num{10000}$ trying up to \texttt{n.threads=6}, so we do not enable
		parallelization in \texttt{dbarts}. Similarly, we don't
		use the data-sharding capability of \texttt{bartz} across
		virtual CPU devices, which can make it faster by up to 2x.

	\item We did not find a reliable way to set the number of cores used by
		JAX on macOS, so for \texttt{bartz} we observed rather than
		imposed that it never uses beyond the 6 high-performance cores
		of the M1 Pro, with average load around 2.

	\item XGBoost and CatBoost both have working number-of-threads
		settings, so we set both to 6.
	
	\item We claim \texttt{dbarts} is the previous fastest BART
		implementation. This is mentioned in the literature in
		\textcite[\S1]{he2019} and \textcite[\S D.3,
		p.~61]{sparapani2021}, but the fact that this is still true at
		the time of writing in 2026 is based on unreported
		measurements.
	
	\item We change some \texttt{dbarts} settings to make it match
		\texttt{bartz}: \texttt{n.cuts=255}, which should not have an
		effect on running time, and \texttt{keepTrainingFits=FALSE},
		which should speed up \texttt{dbarts} a tiny bit.
	
	\item We recorded the results for CatBoost on CPU, but they
		are not displayed in the plots to reduce clutter. In these
		benchmarks, CatBoost mostly serves as a reference to check that
		XGBoost is working fine.
	
	\item Each power-of-2 $n$ sweep terminates at the first $n$ where the
		runner hits a 120\,s timeout. When running \texttt{bartz} on
		GPU, it can also stop if JAX throws an out-of-memory error.

\end{itemize}

\section{Details of comparison between BART packages}
\label{sec:compdetails}

\begin{figure}
	\centering
	\widecenter{\includempl{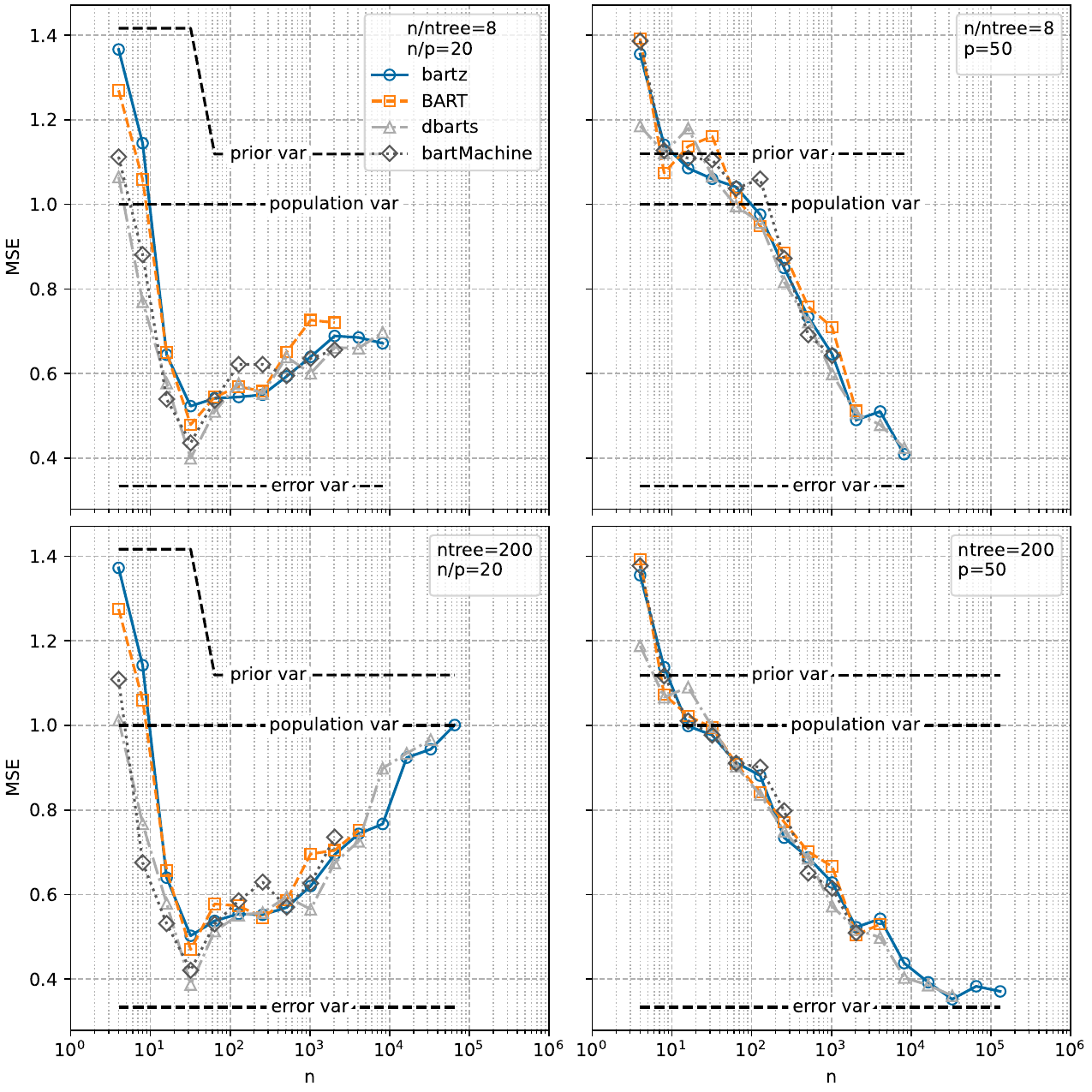}}
	\caption{\label{fig:rmse-all} Full version of
		\autoref{fig:accuracy-single}, left panel. Test set mean squared
		error (MSE) vs.\ training set size~$n$, comparing
		\texttt{bartz} with the most popular implementations of BART,
		to check they yield the same predictive performance, for
		various settings of number of trees and number of predictors
	$p$ on data simulated as \autoref{eq:dgp}. The test set size is always
1000. The MSE is averaged over enough runs to reach a total training set size
of at least 1000 for each point, capped at 10 runs. The data is re-generated independently for each method.}
\end{figure}

\begin{figure}
	\centering
	\widecenter{\includempl{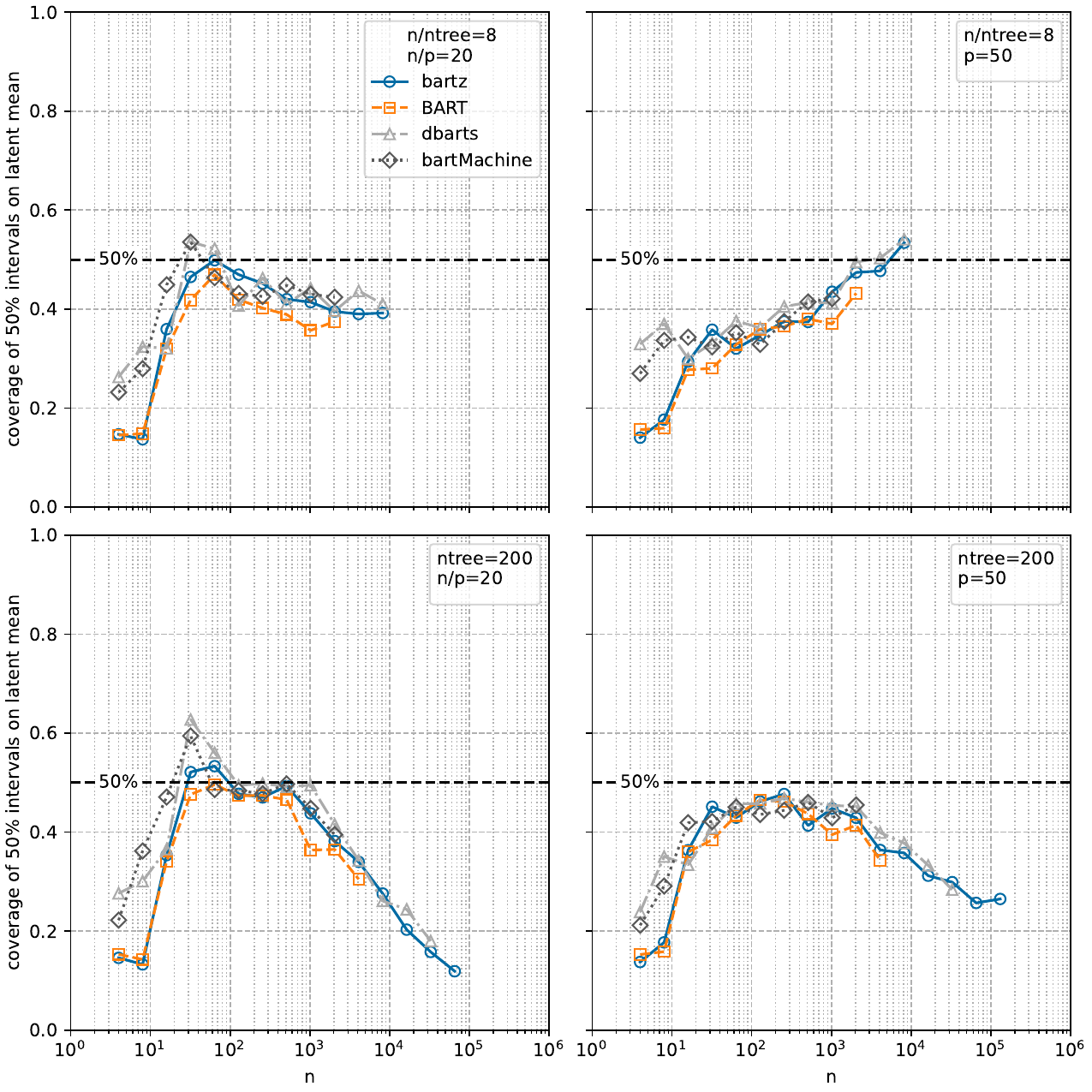}}
	\caption{\label{fig:coverage-all} Full version of
		\autoref{fig:accuracy-single}, right panel. Coverage of
		\SI{50}{\%} credible intervals on the latent mean vs.\ training
		set size~$n$, comparing \texttt{bartz} with the most popular
		implementations of BART, to check they yield the same
		predictive performance, for various settings of number of trees
		and number of predictors $p$ on data simulated as
		\autoref{eq:dgp}. The test set size is always 1000. Coverage is
		averaged over enough runs to reach a total training set size of
	at least 1000 for each point, capped at 10 runs. The data is re-generated independently for each method.}
\end{figure}

The full results of the predictive accuracy comparison between BART packages
are in Figures~\ref{fig:rmse-all} and~\ref{fig:coverage-all}. Some details:
\begin{itemize}

	\item Here for \texttt{BART} we actually use the \texttt{BART3} package
		\autocite{mcculloch2025}, its not formally released successor,
		and we use \texttt{bartz} through its compatibility interface
		\texttt{bartz.BART.gbart} meant to closely reproduce the
		specific settings of \texttt{BART3}.
	
	\item All runs are on CPU. The power-of-2 $n$ sweep terminates,
		separately for each method, when the runner hits a 120\,s
		timeout, so slower packages are tested up to smaller $n$.
	
	\item We change the settings of the packages to make them as consistent
		as possible with each other. Some irreducible differences
		remain, for example \texttt{bartMachine} does not allow
		controlling the grid of splitting points.
	
	\item In \autoref{fig:rmse-all}, the dashed lines labelled
		``prior/population/error var'' refer to the quantities defined
		in Equations~\ref{eq:popvar}, \ref{eq:priorvar}, and
		$\sigma^2_\text{eps}$ in \autoref{eq:dgp}.

\end{itemize}

\section{Details of the benchmark on datasets}
\label{sec:fulldetails}

The full results of the \texttt{bartz} vs.\ XGBoost comparison on a few
datasets are in \autoref{tab:fullbench-full}.

\begin{table}
	\centering
	\caption{\label{tab:fullbench-full} Extended version of
		\autoref{tab:fullbench}. Comparison of \texttt{bartz} against
		XGBoost on various large datasets, at near-default
		settings, on an Nvidia RTX PRO 5000 GPU. Each run is repeated 5
		times over different train/test splits and with different
	random seeds. The standard deviation across repetitions is reported as
a digit in parentheses when non-zero. See \autoref{sec:fulldetails} for
details.}
	\footnotesize\sffamily
	\renewcommand{\arraystretch}{0.97}
	% typeset the numbers as text, in the current font family
	\sisetup{mode=text, reset-text-family=false}
	\setlength{\tabcolsep}{3pt}
	\widecenter{\begin{tabular}{llrrrrrr}
\toprule
 &  & electricity & higgs & \begin{tabular}[b]{@{}r@{}}radar \\ traffic\end{tabular} & \begin{tabular}[b]{@{}r@{}}zurich \\ delays\end{tabular} & poker & \begin{tabular}[b]{@{}r@{}}simulated \\ (\autoref{eq:dgp})\end{tabular} \\
\midrule
OpenML id &  & \href{https://www.openml.org/d/46237}{46237} & \href{https://www.openml.org/d/45570}{45570} & \href{https://www.openml.org/d/43354}{43354} & \href{https://www.openml.org/d/45045}{45045} & \href{https://www.openml.org/d/43955}{43955} & --- \\
\textit{n} &  & \num{1145264} & \num{10990000} & \num{1958768} & \num{5455575} & \num{1012616} & \num{3990000} \\
\textit{p} &  & \num{11} & \num{28} & \num{14} & \num{20} & \num{5} & \num{1000} \\
task &  & regr. & class. & regr. & regr. & class. & regr. \\
simulated &  & no & yes & no & no & yes & yes \\
time series &  & yes & no & yes & yes & no & no \\
test sdev &  & \num{0.998(4)} & \num{0.4993(2)} & \num{1.002(3)} & \num{0.997(7)} & \num{0.49998(2)} & \num{1.047(4)} \\
\midrule
RMSE & xgboost & \num{0.136(2)} & \num{0.414(3)} & \num{0.258(2)} & \num{0.946(7)} & \num{0.32(1)} & \num{0.707(3)} \\
 & bartz & \num{0.138(2)} & \num{0.417(2)} & \num{0.272(1)} & \num{0.946(7)} & \num{0.116(6)} & \num{0.790(4)} \\
 & xgboost-es & \num{0.119(2)} & \num{0.397(2)} & \num{0.208(5)} & \num{0.942(6)} & \num{0.044(5)} & \num{0.621(2)} \\
 & bartz-2000 & \num{0.131(2)} & \num{0.403(2)} & \num{0.227(3)} & \num{0.945(7)} & \num{0.076(3)} & \num{0.601(3)} \\
\addlinespace
error sdev & bartz & \num{0.1374(2)} & --- & \num{0.273(1)} & \num{0.94640(3)} & --- & \num{0.790(2)} \\
 & bartz-2000 & \num{0.1288(1)} & --- & \num{0.2242(2)} & \num{0.94479(3)} & --- & \num{0.6029(9)} \\
\addlinespace
MSE/test var & xgboost & \num{0.0186(5)} & \num{0.688(9)} & \num{0.066(1)} & \num{0.900(2)} & \num{0.42(3)} & \num{0.456(4)} \\
 & bartz & \num{0.0192(5)} & \num{0.697(8)} & \num{0.0739(3)} & \num{0.9002(8)} & \num{0.053(5)} & \num{0.569(9)} \\
 & xgboost-es & \num{0.0141(4)} & \num{0.633(5)} & \num{0.043(2)} & \num{0.894(2)} & \num{0.008(2)} & \num{0.352(3)} \\
 & bartz-2000 & \num{0.0171(5)} & \num{0.652(8)} & \num{0.051(1)} & \num{0.898(2)} & \num{0.023(2)} & \num{0.329(3)} \\
\addlinespace
log loss & xgboost & --- & \num{0.513(6)} & --- & --- & \num{0.35(2)} & --- \\
 & bartz & \num{-0.56(1)} & \num{0.520(4)} & \num{0.117(5)} & \num{1.363(7)} & \num{0.073(5)} & \num{1.183(6)} \\
 & xgboost-es & --- & \num{0.478(4)} & --- & --- & \num{0.015(3)} & --- \\
 & bartz-2000 & \num{-0.62(1)} & \num{0.490(5)} & \num{-0.07(1)} & \num{1.362(7)} & \num{0.047(2)} & \num{0.910(5)} \\
\addlinespace
50\,\% coverage & bartz & \num{0.685(7)} & --- & \num{0.658(6)} & \num{0.462(7)} & --- & \num{0.501(7)} \\
 & bartz-2000 & \num{0.681(4)} & --- & \num{0.684(2)} & \num{0.462(8)} & --- & \num{0.505(5)} \\
\addlinespace
train time (s) & xgboost & \num{0.38(4)} & \num{1.62(3)} & \num{0.46(2)} & \num{0.72(3)} & \num{0.32(3)} & \num{11.09(5)} \\
(incl. XLA) & bartz & \num{19.24(5)} & \num{74.1(4)} & \num{24.53(8)} & \num{59.6(3)} & \num{18.35(3)} & \num{42.6(6)} \\
 & xgboost-es & \num{2.7(3)} & \num{37(3)} & \num{9(1)} & \num{7.8(6)} & \num{1.88(7)} & \num{51(5)} \\
(incl. XLA) & bartz-2000 & \num{96.14(4)} & \num{537(1)} & \num{134.87(9)} & \num{332.5(2)} & \num{96.80(6)} & \num{223.4(3)} \\
\addlinespace
cold predict (s) & xgboost & \num{0.00260(2)} & \num{0.0040(8)} & \num{0.00270(5)} & \num{0.0038(8)} & \num{0.00258(4)} & \num{0.0309(4)} \\
(incl. XLA) & bartz & \num{1.33(1)} & \num{1.07(1)} & \num{1.328(9)} & \num{1.338(8)} & \num{1.04(1)} & \num{1.59(1)} \\
 & xgboost-es & \num{0.021(3)} & \num{0.048(3)} & \num{0.06(1)} & \num{0.025(2)} & \num{0.0184(7)} & \num{0.040(1)} \\
(incl. XLA) & bartz-2000 & \num{0.980(4)} & \num{1.112(5)} & \num{1.00(3)} & \num{0.984(4)} & \num{1.091(2)} & \num{2.76(3)} \\
\addlinespace
warm predict (s) & xgboost & \num{0.00096(4)} & \num{0.00120(7)} & \num{0.00096(3)} & \num{0.00096(6)} & \num{0.00088(1)} & \num{0.0285(3)} \\
 & bartz & \num{0.1889(2)} & \num{0.198(1)} & \num{0.1930(4)} & \num{0.1941(6)} & \num{0.1874(2)} & \num{0.4097(4)} \\
 & xgboost-es & \num{0.0051(5)} & \num{0.0113(8)} & \num{0.014(2)} & \num{0.0059(5)} & \num{0.0043(1)} & \num{0.033(4)} \\
 & bartz-2000 & \num{0.2049(6)} & \num{0.2143(3)} & \num{0.2103(3)} & \num{0.2057(2)} & \num{0.1983(5)} & \num{1.877(4)} \\
\addlinespace
peak GPU mem. & xgboost & \num{0.593(8)} & \num{2.82} & \num{0.9(1)} & \num{2.00} & \num{0.560} & \num{24.5} \\
(GB) & bartz & \num{0.848} & \num{3.19} & \num{0.973} & \num{1.80} & \num{0.835} & \num{22.0} \\
 & xgboost-es & \num{0.60(4)} & \num{2.69} & \num{0.8(2)} & \num{1.95} & \num{0.566} & \num{22.1} \\
 & bartz-2000 & \num{3.56} & \num{23.8} & \num{5.32} & \num{12.3} & \num{3.42} & \num{22.0} \\
\addlinespace
trees & xgboost & \num{100} & \num{100} & \num{100} & \num{100} & \num{100} & \num{100} \\
 & bartz & \num{200} & \num{200} & \num{200} & \num{200} & \num{200} & \num{200} \\
 & xgboost-es & \num{1617(190)} & \num{3672(287)} & \num{4495(808)} & \num{1797(151)} & \num{1353(51)} & \num{724(84)} \\
 & bartz-2000 & \num{2000} & \num{2000} & \num{2000} & \num{2000} & \num{2000} & \num{2000} \\
\addlinespace
leaves/64 & xgboost & \num{0.9955(8)} & \num{0.9982(9)} & \num{0.998(1)} & \num{0.968(3)} & \num{0.80(5)} & \num{1.00} \\
\phantom{leaves}/32 & bartz & \num{0.51(1)} & \num{0.895(5)} & \num{0.929(8)} & \num{0.352(6)} & \num{0.492(5)} & \num{0.56(1)} \\
\phantom{leaves}/64 & xgboost-es & \num{0.9909(2)} & \num{0.9868(5)} & \num{0.9927(4)} & \num{0.956(2)} & \num{0.893(3)} & \num{0.9979(2)} \\
\phantom{leaves}/32 & bartz-2000 & \num{0.166(1)} & \num{0.321(1)} & \num{0.307(2)} & \num{0.1151(5)} & \num{0.205(2)} & \num{0.0986(7)} \\
\addlinespace
empty trees/0.05 & bartz & \num{0} & \num{0} & \num{0} & \num{0.02(3)} & \num{0} & \num{0} \\
 & bartz-2000 & \num{0.77(5)} & \num{0.060(6)} & \num{0.08(1)} & \num{0.73(2)} & \num{0.062(8)} & \num{0.59(6)} \\
\addlinespace
acceptance & bartz & \num{0.027(1)} & \num{0.0124(6)} & \num{0.0119(8)} & \num{0.054(1)} & \num{0.0064(3)} & \num{0.0219(4)} \\
 & bartz-2000 & \num{0.0637(8)} & \num{0.0528(4)} & \num{0.0462(6)} & \num{0.125(1)} & \num{0.0420(4)} & \num{0.0400(9)} \\
\addlinespace
tree life/2000 & bartz & \num{0.99978(8)} & \num{0.99980(5)} & \num{0.99990(3)} & \num{0.986(6)} & \num{0.9992(2)} & \num{0.9961(7)} \\
 & bartz-2000 & \num{0.802(8)} & \num{0.945(4)} & \num{0.962(4)} & \num{0.573(5)} & \num{0.917(5)} & \num{0.76(1)} \\
\bottomrule
\end{tabular}
}
\end{table}

\paragraph{XGBoost setup} 

We use the main XGBoost Python interface, \texttt{xgboost.train()}, but
imitating the settings of the high-level scikit-learn-like interfaces
\texttt{xgboost.XGBRegressor} and \texttt{xgboost.XGBClassifier}. The only change we make is that we move the data
to GPU using CuPy \autocite{cupy_learningsys2017} before passing it into \texttt{xgboost.QuantileDMatrix}, and
instead of using \texttt{cupy.array(...)} as in the XGBoost GPU tutorial, we
use \texttt{x = cupy.empty(...); x.set(...)} because we found this to be much
faster. With the default \texttt{XGBRegressor}-like code path, the predictors would
be pre-processed on CPU. For ``xgboost-es'', we set aside a random 10\,\% of the
training split as validation set. At every round (i.e., after generating each
tree), XGBoost will stop if the last 50 rounds showed no loss decrease on the
validation set, and we truncate the trees at the best one (so 50 before the last
one produced).

\paragraph{bartz setup}

We use the high-level interface \texttt{bartz.Bart}, changing only the number
of chains from the default 4 to 1. With small datasets, using more chains would
be near-free on GPU due to parallelization, while with the large datasets of
this benchmark, the dataset itself already saturates the GPU. There is possibly
one unfair advantage compared to the XGBoost setup, in that CUDA initialization
probably happens outside the timed region for \texttt{bartz}, but inside it for
XGBoost. This may be relevant on smaller datasets without early stopping, where
the XGBoost training time is below 1\,s. We haven't investigated further.

\paragraph{Dataset selection and preprocessing}

The datasets used in \autoref{tab:fullbench-full} were chosen by considering
OpenML datasets with $n\ge\num{1000000}$ and then filtering out those that
appeared low quality on a subjective judgement, were not apt for a
\texttt{bartz} run without nontrivial pre-processing (mainly, \texttt{bartz}
does not support missing predictors and high-cardinality categorical
predictors), or were not apt for a simple generic out-of-sample predictive
comparison. A pre-selection of 27 datasets was finalized before preparing the
benchmark. Further pre-processing and inclusion decisions were made based on general considerations like signal-to-noise ratio without taking into account \texttt{bartz} vs.{} XGBoost comparisons, except for the dataset simulated from
\autoref{eq:dgp}, which was added to have at least one high-$p$ dataset, with full awareness that BART would probably do better because of that, though only a priori. The hyperparameters of the DGP were coarsely tuned to have an out-of-sample MSE around 0.5.

The train/test split is random and does not take into account temporal ordering
on time series data. Thus, on datasets with a time axis, predictions are
interpolations rather than forward predictions.

We mostly followed the pre-processing of the paper introducing the dataset
where we found one: \textcite{grinsztajn_why_2022} for ``poker'' and ``zurich
delays'', \textcite{baldi_searching_2014} for ``higgs''. We presume the full
chain of provenance of our ``poker'' dataset is: UCI Poker Hand dataset \autocite{poker_hand_158}, LibSVM
datasets, OpenML id 354, \textcite{grinsztajn_why_2022}'s pipeline, OpenML id
43955. The ``radar traffic'' dataset comes from public data
\autocite{austin2019radar}; ``electricity'' comes from \textcite{godahewa_2021_4659727}. For the full details of our pre-processing, see the
supplemental code (\autoref{sec:code}).

\paragraph{Definition of measurements reported in the table}

The meaning of the rows in \autoref{tab:fullbench-full} is:
\begin{description}
	
	\item[OpenML id] The unique identifier of the dataset in the OpenML
		catalogue.

	\item[\textit{n}] The number of units in the training split. The test set is
		always \num{10000} units.

	\item[\textit{p}] The number of predictors, after pre-processing.

	\item[task] Whether the outcome is continuous (``regression'') or
		binary (``classification'').
	
	\item[simulated] Whether the data is collected from real-world
		measurements or simulated.

	\item[time series] Whether the data has temporal ordering (we ignore
		it).

	\item[test sdev] The standard deviation of the test set. Always close
		to 1.000 for continuous outcomes because they are standardized,
		while for binary outcomes it indicates balancing. Both our
		binary datasets are balanced.

	\item[RMSE] The root mean squared error of out-of-sample predictions on
		the test set. The MSE is averaged over 5 independent train/test
		splits.

	\item[error sdev] The posterior RMS error standard deviation of BART.

	\item[MSE/test var] Fraction of the variance unexplained by the
		predictions. Reports the same information as the RMSE, but
		ranges from 0 to 1.

	\item[log loss] The negative logarithm of the probability (density) assigned to the
		correct test label (outcome) by the model, averaged over the test set.
		For XGBoost, this is available only for classification.

	\item[50\,\% coverage] BART-only. Coverage of the \SI{50}{\%} credible
		intervals on test outcomes, when outcomes are continuous.

	\item[train time] The end-to-end time to train the model on a single
		split, starting cold. Each method handles copying the data from
		CPU to GPU and pre-processing it, included in this time.
	
	\item[cold predict] The time taken to compute predictions and scores given a
		trained model, on the first prediction after training.

	\item[warm predict] The time taken to compute predictions and scores given a
		trained model, re-computing them a second time.

	\item[peak GPU mem.] Maximum GPU memory used during a full run
		(loading, training, prediction, scoring). For XGBoost, this
		measures the total allocated GPU memory. This is not possible
		with \texttt{bartz} because JAX aggressively keeps memory ownership,
		so we query the JAX allocator instead.
		This is an underestimate by probably about
		\SI{1}{GB} because there is additional memory usage beyond JAX
		arrays. The figures for \texttt{bartz} have no variation because
		they were measured on a single run instead of 5 times.

	\item[trees] The number of trees.

	\item[leaves] The average number of leaves per tree (averaged over
		post-burn-in MCMC samples for BART), reported as a fraction of
		the total number of leaves allowed by the method.

	\item[empty trees] BART-only. The fraction of empty (i.e., root-only,
		depth-0) trees across post-burn-in MCMC samples, reported as a
		ratio to the a-priori expected fraction. This ratio is
		typically below 1, but it can go above 1 if there are more
		trees than needed to explain the signal.
	
	\item[acceptance] BART-only. The average acceptance of the tree
		structure Metropolis-Hastings in the post-burn-in MCMC samples.

	\item[tree life] BART-only. The number of iterations since the last
		time the tree was pruned back to empty, averaged over trees at
		the last MCMC iteration, and reported as a fraction of the
		total number of iterations.

\end{description}

\section{Comment on the constraint of GPU memory}
\label{sec:gpumem}

A striking feature of the speed benchmarks in \autoref{fig:time-all} is that
with high $p$ or $n_\text{tree}$ the \texttt{bartz-P5000} line terminates at
small $n$ like the other methods despite being faster. Our benchmark harness
increases $n$ until timeout or out-of-memory error, so this indicates that
the benchmark is running out of GPU memory. The memory installed on GPU cards
is fixed, it can't be expanded, and there is no swap, so this is a hard limit.
The speed premium of the GPU kicks in only at high $n$, so this memory limit
indirectly limits the speed gain. The practical consequence is that, with 48
GiB of GPU memory, running the algorithm in ``high'' $p$ or $n_\text{tree}$
settings allows a GPU over CPU speed gain of at most about 25--50x (extrapolating the CPU lines) rather than the
asymptotic 100x.
Some of this limitation comes from our benchmark setup, which incidentally
limits memory usage for JAX to 75\,\% of the total, and in high~$p$ squats a large fraction of this memory for
generating the simulated data. However, fixing these issues would not change
the fundamental constraint.

\section{On increasing the number of trees}
\label{sec:moretrees}

Here we discuss how the speed of this implementation allows running BART with many
more trees than usual, which is a little-explored route in the BART literature, and
potentially connects to various scaling behaviors of BART. Our a priori best
guess is that, since 1) the total number of leaves in the set of trees is the
effective number of parameters of the regression model, 2) the optimal number
of parameters will in general increase with $p$ and $n$, and 3) the trees ought
to remain shallow, then: the number of trees should increase up to linearly
with $\max(n, p)$.

We look first at our MSE measurements to get some empirical evidence that
increasing the number of trees is useful. Consider the left column of panels in
\autoref{fig:rmse-all}. They both have a ``high'' $p$ setting ($n = 20p$),
while differing in the number of trees. In the bottom panel, at fixed
$n_\text{tree}=200$, the MSE increases with $n$, reaching at some point the
population variance, indicating useless predictions; while in the top panel,
with $n_\text{tree}=n/8$, the MSE appears to be following a flat trend (a
complete direct comparison is not possible due to the different truncation of
the plots due to memory and time constraints in the runs). Analogous
considerations apply to the left column in \autoref{fig:coverage-all}. Next,
consider the \texttt{bartz} vs.{} XGBoost comparison in \autoref{tab:fullbench-full}:
``bartz-2000'' (2000 trees) dominates ``bartz'' (200 trees) on RMSE and MCMC
mixing (``empty trees'', ``acceptance'', ``tree life'').

The number of trees bears on the acceptance of the BART MCMC, together with the
level of noise in the data. If the number of trees is too small, each tree has
to grow deep to do its part in explaining the patterns in the data, and the MCMC
gets stuck because ungrowing the trees is disfavored, and so they can't be
changed in little steps; while with enough trees, each tree has to explain only
a small part of the variation, so the prior prevails and keeps them shallow. The
level of noise (i.e., unpredictable variation) has a similar effect
\citep{pratola2016}: if the data is precisely predictable (for concreteness,
picture $\operatorname{Std}[\mathbf y]/\sigma = 100$), the trees quickly grow to
explain the variation, the error variance is shrunk accordingly, and once it is
small, any small change that doesn't preserve a nigh-perfect data fit is too
disfavored, so the MCMC gets stuck. Thus the signal/noise ratio (SNR) is an
important influence on the optimal number of trees; high SNR datasets may need
more trees.

\textcite[pp.~61--64]{collins_improved_2023} reaches similar conclusions: the
number of trees should in general increase with number of datapoints, SNR, and
complexity of the signal.

Our work pushes the number of trees farther past everything else we found in
the literature, which mostly uses the common 200 trees default. We run BART with up to 1024 trees in
Figures~\ref{fig:rmse-all}, \ref{fig:coverage-all} and up to 2000 in
\autoref{tab:fullbench-full}. The example notebook we provide with the
documentation of \texttt{bartz} uses \num{10000} trees.
The other works using the next highest number of trees, that we know of,
are \textcites
[][\S5.3, p.~289, 800 trees]{chipman2010}
[][\S7.3, p.~849, 500 trees]{pratola2014}
[][Fig.~3.4, p.~57, 800 trees]{collins_improved_2023}
[][Fig.~6, p.~26, 500 trees]{tan2026computationalefficiencybayesianadditive}.

Consider these examples from the literature, where it appears the
authors were not considering increasing the number of trees as an option:

\begin{itemize}

	\item \citet[\S4.1]{ronen2022} (the precursor of
		\cite{tan2026computationalefficiencybayesianadditive}) showed
		numerically how the convergence of the MCMC degrades at high
		$n$, but didn't consider increasing the number of trees. The
		2024 version of
		\textcite{tan2026computationalefficiencybayesianadditive}
		suggested increasing the number of trees, but did not seem to
		consider it as interesting as other options, and only tried up
		to 10~trees in experiments on varying the number of trees. Then at last, the 2026 revision
		frames increasing the number of trees as an important knob for MCMC
		mixing, and runs experiments with up to 500 trees.

	\item \citet{pratola2016} develops complex Metropolis proposals to increase
	      acceptance in high-$n$, low-$\sigma$ datasets; a simpler alternative
	      solution might be again to increase the number of trees.

	\item \citet{lakshminarayanan_particle_2015} introduce an alternative BART
	      MCMC algorithm, claiming that it addresses scaling problems of the standard
	      BART MCMC. However, in their empirical comparisons (\S4), to fit a
	      high-dimensional dataset, they set the BART hyperparameters to make the
	      trees grow very deep, or keep the number of trees fixed on complex datasets, instead of simply increasing the number of trees.

\end{itemize}
We hypothesize that the inconvenience of running BART with many trees, simply
due to running time, may have slowed progress in this area.

\section{Code and data}
\label{sec:code}

The results in this paper are fully reproducible with
\redactedcite{\citet{gattocrucco_2026_24033756}}. We use the following open-source software
which is not mentioned elsewhere in the paper: Matplotlib \autocite{matplotlib},
SciPy \autocite{scipy}, polars \autocite{ritchie_vink_2026_19699949},
pandas \autocite{mckinney-proc-scipy-2010}, R \autocite{erre}, rpy2
\autocite{gautier2026rpy2}.

\end{document}